\documentclass[10pt,journal,compsoc]{IEEEtran}
\usepackage{times}
\usepackage{epsfig}
\usepackage{threeparttable}
\usepackage{graphicx}
\usepackage{amsmath}
\usepackage{amssymb}
\usepackage{float}
\usepackage{booktabs}
\usepackage{multirow}
\usepackage{rotating}
\usepackage{bbding}
\usepackage{enumitem}
\usepackage{bm}
\usepackage{ragged2e} 
\usepackage{array}
\usepackage{math_symbols}
\usepackage[pagebackref=false,breaklinks=true,colorlinks,bookmarks=false]{hyperref}
\usepackage[ruled,vlined]{algorithm2e}
\usepackage[font=small]{caption}
\usepackage{graphicx}
\usepackage[usenames,dvipsnames]{xcolor}
\usepackage{pdfpages}
\usepackage{colortbl} 
\usepackage{footmisc}

\definecolor{mypink}{cmyk}{0, 0.7808, 0.4429, 0.1412}
\definecolor{mygreen}{rgb}{0.0, 0.7, 0.0}
\definecolor{myblue}{rgb}{0.0, 0.72, 0.92}
\definecolor{mygray}{gray}{0.6}
\definecolor{mygray-bg}{gray}{0.9}

\newcommand{\ie}{\textit{i}.\textit{e}.}
\newcommand{\eg}{\textit{e}.\textit{g}.}
\newcommand{\cf}{\textit{\cf,} }
\newcommand{\etal}{\textit{et}.\textit{al}.}

\newcommand{\vs}{\textit{vs}. }

\newcommand{\shr}[1]{{\color{black}{#1}}}

\newcommand{\sgrouptablestyle}[2]{\setlength{\tabcolsep}{#1}\renewcommand{\arraystretch}{#2}\centering}
\newcolumntype{x}[1]{>{\centering\arraybackslash}p{#1pt}}
\newcolumntype{I}{!{\vrule width 1pt}}
\newcolumntype{d}[1]{>{\raggedright\arraybackslash}p{#1pt}}
\newcolumntype{b}[1]{>{\raggedleft\arraybackslash}p{#1pt}}

\newcommand{\subt}[4]{
    \sgrouptablestyle{0pt}{1}
    \begin{tabular}{b{#1}d{#2}}
    {#3} & {#4}
    \end{tabular}
}

\makeatletter
\DeclareRobustCommand\onedot{\futurelet\@let@token\@onedot}
\def\@onedot{\ifx\@let@token.\else.\null\fi\xspace}

\def\eg{\textit{e.g}\onedot} 
\def\ie{\textit{i.e}\onedot} 
\def\cf{\textit{c.f}\onedot} 
 \def\vs{\textit{vs}\onedot}
 
\def\etal{\textit{et al}\onedot}
\def\cf{\textit{cf}\onedot}
\makeatother

\makeatletter
\newcommand{\thickhline}{%
    \noalign {\ifnum 0=`}\fi \hrule height 1pt
    \futurelet \reserved@a \@xhline
}

\makeatother




\usepackage{graphicx} 
\makeatother
\ifCLASSOPTIONcompsoc
  \usepackage[nocompress]{cite}
\else
  \usepackage{cite}
\fi
\ifCLASSINFOpdf
\else
\fi
\ifCLASSOPTIONcompsoc
 \usepackage[caption=false,font=footnotesize,labelfont=sf,textfont=sf]{subfig}
\else
 \usepackage[caption=false,font=footnotesize]{subfig}
\fi
\begin{document}
%
\title{NICEST: Noisy Label Correction and Training for Robust Scene Graph Generation}
%
%
%
%

\author{Lin~Li,
        Jun~Xiao,
        Hanrong~Shi,
        Hanwang~Zhang,
        Yi~Yang,
        Wei~Liu,
        and~Long~Chen$^*$
\IEEEcompsocitemizethanks{
        \IEEEcompsocthanksitem $^*$Long Chen is the corresponding author.
		\IEEEcompsocthanksitem Lin Li, Jun Xiao, Hanrong Shi, and Yi Yang are with the College of Computer Science, Zhejiang University, Hangzhou, 310027, China. E-mails: \{mukti, junx, hanrong, yangyics\}@zju.edu.cn. 
		\IEEEcompsocthanksitem Hanwang Zhang is with the School of Computer Science and Engineering, Nanyang Technological University, 639798, Singapore.	E-mails: hanwangzhang@ntu.edu.sg.
		\IEEEcompsocthanksitem Wei Liu is with the Data Platform, Tencent, Shenzhen, 518000, China. Email: wl2223@columbia.edu.
            \IEEEcompsocthanksitem Long Chen is with the Department of Computer Science and Engineering, The Hong Kong University of Science and Technology, Hong Kong SAR 999077. E-mail: longchen@ust.hk.
		\IEEEcompsocthanksitem Codes are available: \url{https://github.com/HKUST-LongGroup/NICEST}.
        }
}

%
%

\markboth{IEEE TRANSACTIONS ON PATTERN ANALYSIS AND MACHINE INTELLIGENCE}%
{IEEE TRANSACTIONS ON PATTERN ANALYSIS AND MACHINE INTELLIGENCE}
%



\IEEEtitleabstractindextext{%

\justifying

\begin{abstract}
Nearly all existing scene graph generation (SGG) models have overlooked the ground-truth annotation qualities of mainstream SGG datasets, \ie, they assume: 1) all the manually annotated positive samples are equally correct; 2) all the un-annotated negative samples are absolutely background. In this paper, we argue that neither of the assumptions applies to SGG: there are numerous ``noisy'' ground-truth predicate labels that break these two assumptions and harm the training of unbiased SGG models. To this end, we propose a novel \emph{\textbf{N}o\textbf{I}sy label \textbf{C}orr\textbf{E}ction and \textbf{S}ample \textbf{T}raining} strategy for SGG: \textbf{NICEST}, which rules out these noisy label issues by generating high-quality samples and designing an effective training strategy. Specifically, it consists of: 1) \textbf{NICE}: it detects noisy samples and then reassigns higher-quality soft predicate labels to them. To achieve this goal, NICE contains three main steps: negative Noisy Sample Detection (Neg-NSD), positive NSD (Pos-NSD), and Noisy Sample Correction (NSC). Firstly, in Neg-NSD, it is treated as an out-of-distribution detection problem, and the pseudo labels are assigned to all detected noisy negative samples. Then, in Pos-NSD, we use a density-based clustering algorithm to detect noisy positive samples. Lastly, in NSC, we use weighted KNN to reassign more robust soft predicate labels rather than hard labels to all noisy positive samples. 2) \textbf{NIST}: it is a multi-teacher knowledge distillation based training strategy, which enables the model to learn unbiased fusion knowledge. A dynamic trade-off weighting strategy in NIST is designed to penalize the bias of different teachers. Due to the model-agnostic nature of both NICE and NIST, NICEST can be seamlessly incorporated into any SGG architecture to boost its performance on different predicate categories. In addition, \shr{to better assess the generalization ability of SGG models, we propose a new benchmark, \textbf{VG-OOD}, by reorganizing the prevalent VG dataset. This reorganization deliberately makes the predicate distributions between the training and test sets as different as possible for each subject-object category pair.} This new benchmark helps disentangle the influence of subject-object category biases. Extensive ablations and results on different backbones and tasks have attested to the effectiveness and generalization ability of each component of NICEST.
\vspace{-1em}
\end{abstract}


}

\maketitle

\IEEEdisplaynontitleabstractindextext

%
\IEEEpeerreviewmaketitle

\IEEEraisesectionheading{\section{Introduction}\label{sec:introduction}}

\begin{figure*}[t]
    \centering
    \includegraphics[width=\linewidth]{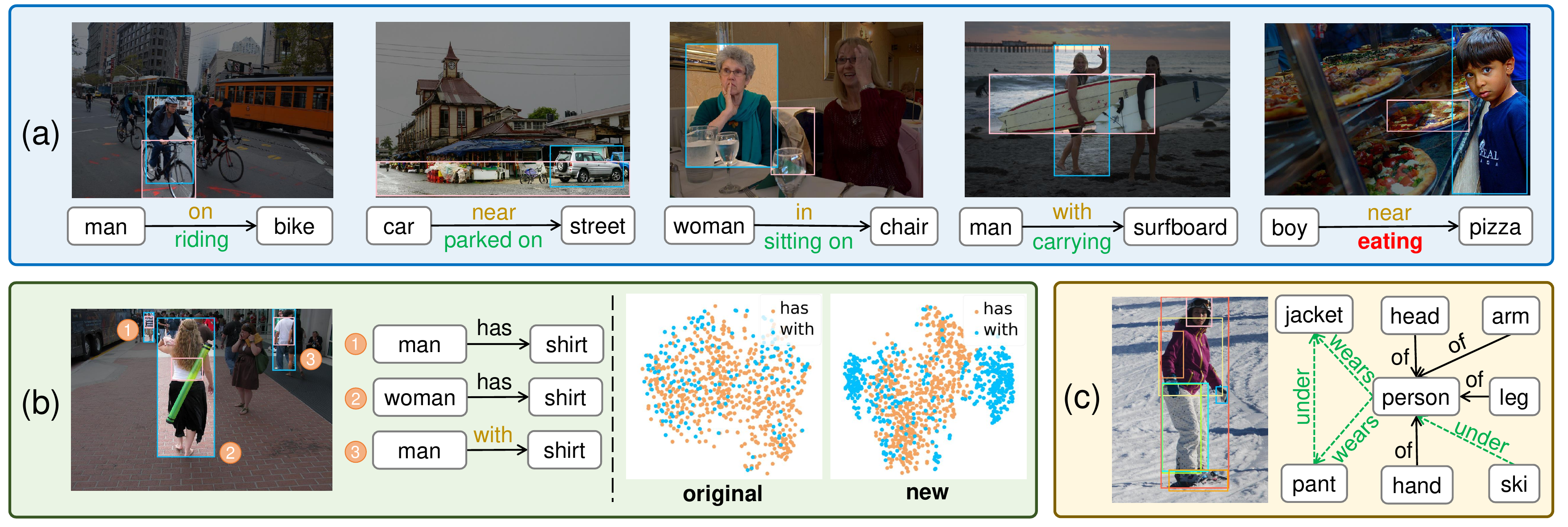}
    \vspace{-0.6em}
    \caption{An illustration of three types of noisy annotations in SGG datasets\shr{, taking VG as an example}. (a) \textbf{Common-prone}: For some triplets, the annotators tend to select less informative coarse-grained predicates (\textcolor{brown}{\textbf{brown}}) instead of the fine-grained ones (\textcolor{mygreen}{\textbf{green}}). The subject and object for each triplet are denoted by \textcolor{myblue}{\textbf{blue}} and \textcolor{mypink}{\textbf{pink}} boxes, respectively. (b) \textbf{Synonym-random}: For some triplets, annotators usually randomly choose one predicate from the several synonyms (\eg, \texttt{has} and \texttt{with} are synonyms for $\langle$\texttt{man/woman}-\texttt{shirt}$\rangle$). \emph{Original:} The $t$-SNE visualization of original triplets $\langle$\texttt{man}-\texttt{has/with}-\texttt{shirt}$\rangle$ features. For brevity, we randomly sample parts of triplets for each type. \emph{New}: The $t$-SNE visualization of same triplets after NICE. (c) \textbf{Negative}: Some negative triplets may not be \texttt{background} (the \textcolor{mygreen}{\textbf{green}} dash arrows).
    }
    \label{fig:motivation}
    \vspace{-0.5em}
\end{figure*}

\IEEEPARstart{S}{cene} Graph Generation (SGG)\cite{xu2017scene, chang2021comprehensive} is a visual task that involves detecting object instances and classifying their pairwise visual relations in an image, which plays a vital role in comprehensive scene understanding~\cite{li2023zero,li2023compositional2}. SGG typically represents scene graphs as visually-grounded graphs, where nodes represent objects and edges represent visual relations. Alternatively, it can be formulated as a set of \texttt{\textless subject-predicate-object\textgreater} triplets. With its structured representation, SGG has contributed to various downstream tasks, such as image retrieval~\cite{johnson2015image, wang2020cross, feng2015semantic}, visual question answering~\cite{lee2019visual, shi2019explainable, chen2020counterfactual}, and image captioning~\cite{kim2021dense,wang2023learning,wang2022explicit}. In recent years, SGG has gained considerable attention~\cite{li2022ppdl,lin2022hl,li2023decomposed, gao2022classification,gao2023compositional} due to the release of large-scale SGG benchmarks (\eg, Visual Genome (VG)~\cite{krishna2017visual} and GQA~\cite{hudson2019gqa}) and advancements in object detection techniques~\cite{hung2020contextual,shao2021improving,shao2022deep}.

Nevertheless, existing SGG benchmarks~\cite{krishna2017visual}, such as VG, suffer from highly skewed long-tailed ground-truth predicate annotations. For example in VG, the number of annotated ground-truth visual relations for the head category ``on'' is 38 times larger than the tail category ``sitting on''. To address this issue, current approaches in SGG~\cite{zellers2018neural,chen2019counterfactual,tang2019learning,lin2020gps,li2023compositional} can be broadly categorized into two strategies. 1) \textit{Re-balancing strategy}, which involves class-aware re-sampling or loss re-weighting to balance the proportions of different predicate categories during training. 2) \textit{Biased-model-based strategy}, which separates unbiased predictions from pretrained biased SGG models~\cite{tang2020unbiased,yu2021cogtree,chiou2021recovering}.

Although these models have dominated on the debiasing SGG metrics (\eg, mean Recall@K), it is worth noting that nearly all existing SGG work has taken two plausible assumptions about the ground-truth annotations for granted:
\begin{itemize}[leftmargin=0.4cm]
    \item[] \textbf{Assumption 1:} \emph{All the manually annotated positive samples are equally correct.}
    
    \item[] \textbf{Assumption 2:} \emph{All the un-annotated negative samples are absolutely background.}
\end{itemize}

For the first assumption, by ``equally'', we mean that the confidence (or quality) of an annotated ground-truth predicate label for each positive sample\footnote{We use ``sample" to represent the triplet instance interchangeably, and we use ``instance" to denote an instance of visual relation triplet. \label{footnote:sample}} is exactly the same, \ie, all the manually annotated positive predicate labels are of high quality. Different from other \shr{closed-set} classification tasks, where each sample corresponds to a unique ground-truth, some specific subject-object pairs in SGG may have multiple \emph{reasonable} predicates, \ie, the semantics of different predicate categories are interdependent to some extent. Inevitably, this phenomenon has resulted in two annotation characteristics in SGG datasets: 1) \textbf{\emph{Common-prone}}: When the semantic granularities of these reasonable visual relations are different, the annotators tend to choose the commonest (or coarse-grained) predicate as the ground-truth label. As shown in Fig.~\ref{fig:motivation}(a), both \texttt{riding} and \texttt{on} are ``reasonable" for the relation between \texttt{man} and \texttt{bike}, but the annotated ground-truth predicate is the less informative \texttt{on} instead of more convincing \texttt{riding}. And this characteristic occurs frequently in the SGG datasets (\cf, more examples in Fig.~\ref{fig:motivation}(a)). 2) \textbf{\emph{Synonym-random}}: When these reasonable visual relations are synonymous for the subject-object pair, the annotators usually randomly select one predicate as the ground-truth label, \ie, the annotations of some similar visual patterns are inconsistent. For example in Fig.~\ref{fig:motivation}(b), both \texttt{has} and \texttt{with} denote meaning ``be dressed in" for \texttt{man}/\texttt{woman} and \texttt{shirt}, but the ground-truth annotations are inconsistent even in the same image. We further visualize thousands of sampled instances\footref{footnote:sample} of $\langle$\texttt{man}-\texttt{has}/\texttt{with}-\texttt{shirt}$\rangle$ in VG, and these instances are all randomly distributed in the feature space (\cf, Fig.~\ref{fig:motivation}(b)). Thus, \emph{we argue that all the positive samples are NOT equally correct, i.e., a part of positive samples are not high-quality --- their ground-truth labels can be more fine-grained (\cf, common-prone) or more consistent (\cf, synonym-random).}

For the second assumption, although almost all the prior SGG work has noticed that ground-truth visual relations in existing datasets are always sparsely identified and annotated~\cite{lu2016visual} (\cf, Fig.~\ref{fig:motivation}(c)), they still train their models by treating all the un-annotated pairs as \texttt{background}, \ie, there is no visual relation between the subject and object pair. By contrast, \emph{we 
contend that all negative samples are NOT absolutely background, \ie, a part of negative samples are not high-quality, that is, they are actually the foreground with missing annotations.}

In this paper, we try to get rid of these two questionable assumptions and propose to reformulate SGG as a noisy label learning problem. Specifically, we propose a novel model-agnostic \emph{\textbf{N}o\textbf{I}sy label \textbf{C}orr\textbf{E}ction and \textbf{S}ample \textbf{T}raining} strategy for SGG, dubbed \textbf{NICEST}. NICEST mitigates the noisy label learning problem from two aspects: generating more high-quality training samples and training models with a more effective mechanism. To achieve this goal, NICEST consists of two parts: \textbf{N}o\textbf{I}sy label \textbf{C}orr\textbf{E}ction (NICE) and \textbf{N}o\textbf{I}sy \textbf{S}ample \textbf{T}raining (NIST). 

\begin{figure}[!t]
  \centering
  \includegraphics[width=0.95\linewidth]{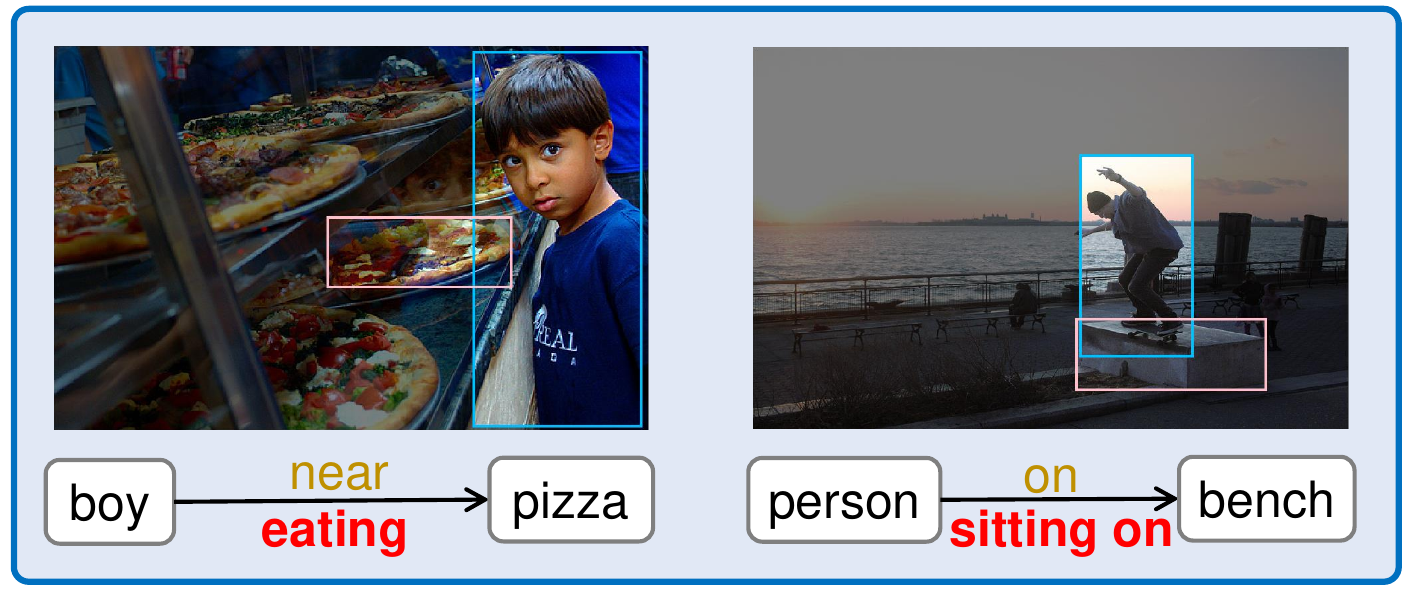}
  \vspace{-0.5em}
  \caption{The triplets wrongly changed by NICE. The original predicates are in \textcolor{brown}{\textbf{brown}} and the changed predicates are in \textcolor{red}{\textbf{red}}.} 
  \label{fig:example_nice}
  \vspace{-0.5em}
\end{figure}

\textbf{NICE.} For high-quality training data, NICE can not only detect numerous \emph{noisy} samples, but also reassign more high-quality predicate labels to them. By ``noisy", we mean that these samples break the two assumptions. After the NICE training, we can obtain a cleaner version of the SGG dataset. Specifically, we can: 1) increase the number of fine-grained predicates (\emph{common-prone}); 2) decrease annotation inconsistency among similar visual patterns (\emph{synonym-random}); 3) increase the number of positive samples (\emph{assumption 2}). 

To be more specific, NICE consists of three steps: 1) negative noisy sample detection (\textbf{Neg-NSD}): We reformulate the negative NSD as an out-of-distribution (OOD) detection problem, \ie, regarding all the positive samples as in-distribution (ID) training data, and all un-annotated negative samples as OOD test data. In this way, we can detect the missing annotated (ID) samples with pseudo labels. 2) positive noisy sample detection (\textbf{Pos-NSD}): We use a clustering-based algorithm to divide all the positive samples (including the outputs of Neg-NSD) into multiple sets, and regard samples in the noisiest set as noisy positive samples. The clustering results are based on the local density of each sample. 3) noisy sample correction (\textbf{NSC}): We utilize the weighted KNN (wKNN) to generate a new pseudo label for each sample. To alleviate the impact of the original noisy label with full probabilities, and give a certain fault tolerance space to the pseudo labels, we reassign soft predicate labels to all noisy positive samples.

\textbf{NIST.} Unfortunately, there is no free lunch, sometimes newly assigned pseudo labels from NICE are not right (\eg, $\langle$\texttt{boy}-\texttt{near}-\texttt{pizza}$\rangle$ is mistakenly changed to $\langle$\texttt{boy}-\texttt{eating}-\texttt{pizza}$\rangle$ in Fig.~\ref{fig:example_nice}). Like other debiasing methods, NICE may over-weight tail categories by increasing the sample percentage of tail categories. For further effective unbiased SGG training, we propose a new knowledge distillation based \textbf{N}o\textbf{I}sy \textbf{S}ample \textbf{T}raining strategy: NIST. NIST is a multi-teacher knowledge distillation method that can integrate the unbiased knowledge of multi-teachers to improve the robustness of distillation. Typically, we can use two different teachers, one is trained on the original dataset (\ie, more in favor of head predicates), and the other is trained on the NICE refined dataset (\ie, more in favor of tail predicates). 

A novel dynamic weighting strategy is devised to punish both biases to obtain unbiased fusion knowledge. Specifically, we first measure the degree of bias of each teacher model by calculating the cross entropy between the predictions of each teacher and the original ground-truth. Then, we fuse the predictions of the low-biased teacher model with the ground-truth. In this way, the biases on different predicates can be eliminated through dynamic weighting. As original ground-truth predicates are proven to be noisy~\cite{li2022devil}, NIST can make the model more robust by smoothing noisy targets to supervise the training. NIST is also a general trade-off training strategy, which can relieve the bias of different predicates for unbiased SGG. More detailed ablations about its generalization ability will be discussed in Sec.~\ref{nist}.

\textbf{New Benchmark VG-OOD.} In addition, for the widely-used SGG benchmarks (\eg, VG~\cite{krishna2017visual}), their predicate distributions of the training set and test set for each subject-object category pair are similar. As displayed in Fig.~\ref{fig:stat}, the predicate distributions of \texttt{woman}-\texttt{shirt} in the original VG dataset are extremely similar (\eg, \texttt{wearing} is the most common predicate in both training and test sets.). Due to this property, a naive baseline that simply uses the frequency priors of the predicates of each subject-object category pair (without considering visual appearance) can achieve satisfactory results~\cite{zellers2018neural}. \shr{For a deeper disentanglement of this prior knowledge and to benchmark the OOD performance of debiasing SGG models}, we propose a new dataset for OOD evaluation: \textbf{VG-OOD}. Specifically, we re-split the VG dataset and try to make the predicate distribution of each subject-object category pair in the training and test sets as inconsistent as possible. In this way, the performance of the test set on VG-OOD can better measure the OOD generalization ability of SGG models.

NICEST was evaluated on the widely recognized scene graph generation (SGG) benchmarks, including VG~\cite{krishna2017visual}, GQA~\cite{hudson2019gqa}, and the newly introduced VG-OOD dataset. Within NICEST, NICE plays a crucial role in enhancing the quality of annotations in original datasets, while NIST focuses on absorbing unbiased knowledge from two teachers. Notably, both NICE and NIST are model-agnostic, allowing for seamless integration into various SGG architectures to enhance their performance. The effectiveness and generalization capabilities of each component within NICEST have been extensively validated through meticulous ablation studies and comprehensive experimental results.

This paper is a substantial and systematic extension of previous CVPR oral work~\cite{li2022devil} with several significant improvements:

\begin{enumerate}[leftmargin=4mm]
    \item In previous NICE-v1\footnote{For the sake of distinction, we use NICE-v1 to denote NICE in~\cite{li2022devil}.}, we merely assign a hard label to noisy samples in NSC, which may introduce new noise. Thus, in this paper, we propose a more robust method to generate soft labels based on the weights in wKNN.
    
    \item Beyond NICE, we propose a noisy label training strategy NIST. It comprehensively considers the unbiased fusion knowledge from multi-teachers and ground-truth labels, which helps achieve robust model training.
    
    \item  We propose a new SGG benchmark for OOD evaluation: VG-OOD. It is designed to supplement the evaluation of the debiasing and generalization ability of SGG models. 
    
    \item We conduct more qualitative and quantitative analyses to demonstrate the effectiveness of NICEST.
\end{enumerate}

\begin{figure*}[!t]
  \centering
  \includegraphics[width=\linewidth]{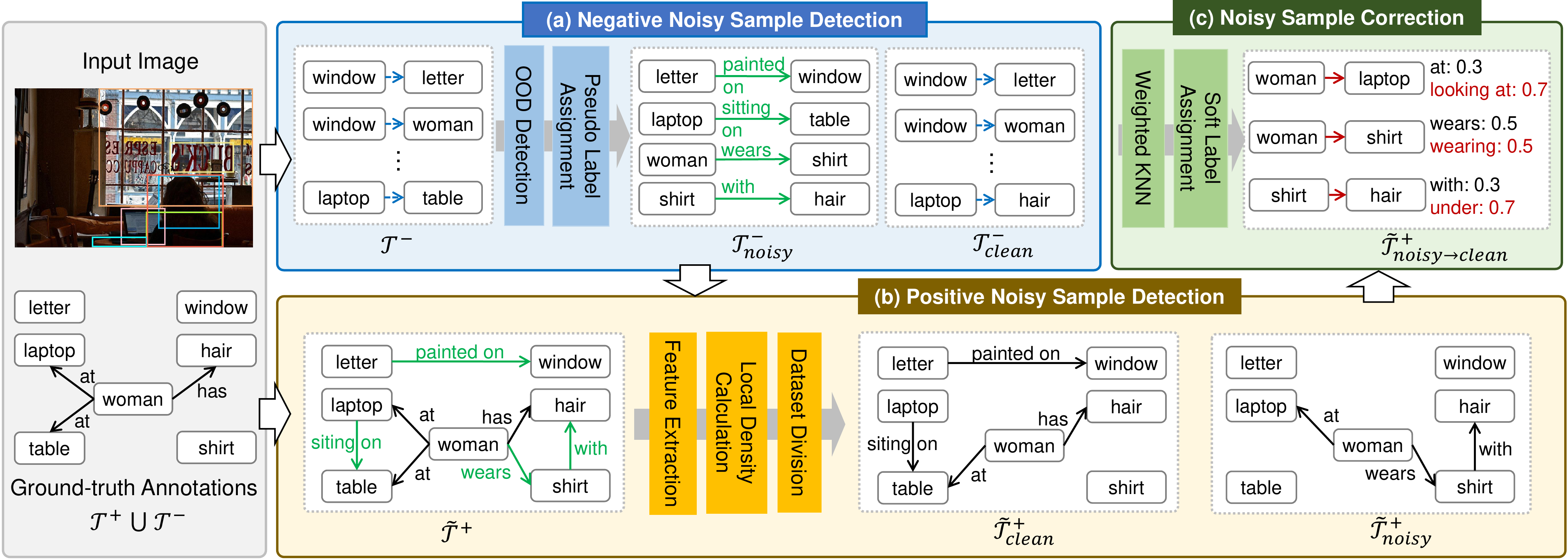}
  \vspace{-1.5em}
  \caption{The pipeline of NICE (taking an image from VG as an example). (a) \textbf{Neg-NSD}: Given all negative triplets (\textcolor{cyan}{\textbf{blue}} dash arrows), the OOD detection model detects missing annotated triplets ($\mathcal{T}^-_{\text{noisy}}$) and assigns pseudo labels to them (\textcolor{mygreen}{\textbf{green}} predicates). (b) \textbf{Pos-NSD}: Given the newly composed positive triplet set ($\widetilde {\mathcal{T}}^+$), Pos-NSD detects all noisy positive samples ($\widetilde {\mathcal{T}}^+_{\text{noisy}}$). (c) \textbf{NSC}: NSC reassigns more high-quality soft predicate labels to all noisy positive samples, and the \textcolor{black}{\textbf{black}} and \textcolor{red}{\textbf{red}} predicates are the scores of the original and raw predicate categories in soft labels. Finally, we obtain a new cleaner version of ground-truth annotations.}
  \label{fig:framework}
\end{figure*}

\section{Related Work}

\textbf{Scene Graph Generation.} SGG helps machines understand visual relationships between objects, which aims to transform the raw visual input into semantic graph structures. Early SGG work~\cite{lu2016visual, zhang2017visual} usually regards each object as an individual and directly predicts their pairwise relations without considering the visual context. Subsequent SGG methods~\cite{xu2017scene, li2017vip, chen2019counterfactual} begin to exploit this overlooked visual context by regarding each image as a whole and adopting the fully connected graph~\cite{yang2018graph,chen2019knowledge}, or the tree-structured graph~\cite{tang2019learning} to model the contexts among objects. Due to the long-tailed distribution and other language bias issues~\cite{misra2016seeing}, unbiased SGG has recently drawn remarkable attention. In general, existing unbiased SGG models can be roughly divided into: re-balancing strategy~\cite{li2021bipartite,lin2017focal,yan2020pcpl,lin2020gps, knyazev2020graph,desai2021learning} and biased-model-based strategy~\cite{tang2020unbiased,yu2021cogtree, chiou2021recovering,guo2021general}. Different from existing work~\cite{li2022label,chen2023addressing}, we are the first to explicitly refine the original noisy ground-truth annotations on SGG datasets, and address the problem in terms of both label quality and training strategy. Although some previous work also has discussed the issue of sparse annotations~\cite{wang2020tackling, chiou2021recovering} or semantic imbalance~\cite{guo2021general}, they still heavily rely on these original noisy annotations in the training.

\textbf{Learning with Noisy Labels.}
Existing noisy label learning methods can be roughly grouped into two categories: 1) Utilizing an explicit or implicit noisy model to estimate the distributions of noisy and clean labels, and then deleting or correcting these noisy samples. These models can be in different formats, such as neural networks~\cite{goldberger2016training, jiang2018mentornet, lee2018cleannet, ren2018learning}, conditional random field~\cite{vahdat2017toward}, or knowledge graphs~\cite{li2017learning}. However, they always need abundant clean training samples, which is inapplicable for many noisy label learning datasets. 2) Constructing a more balanced loss function to reduce the influence of noisy training samples~\cite{liu2015classification,ma2018dimensionality, zhang2018generalized, wang2019symmetric, xu2019l_dmi}. In this paper, we are the first to formulate SGG as a noisy label learning problem, and propose a novel noisy sample correction strategy and an effective training strategy.

\textbf{Knowledge Distillation (KD).} Early KD was proposed to effectively learn lightweight student models from pretrained cumbersome teacher models~\cite{wang2021knowledge, yuan2020revisiting, gou2021knowledge}. Benefiting from the soft targets generated by teacher models, KD has shown potentials in numerous vision tasks, \ie, object detection~\cite{wang2019distilling, chen2017learning,hao2019end}, long-tailed learning~\cite{xiang2020learning, zhang2021balanced, he2021distilling}, scene graph generation~\cite{li2022integrating}, and visual question answering (VQA)~\cite{chen2022rethinking}. However, most traditional KD methods only consider one teacher model. To overcome the limitation of the data diversity and single-teacher knowledge, multi-teacher knowledge distillation~\cite{you2017learning,liu2020adaptive,yang2020model,liang2022multi} integrates comprehensive knowledge of multiple teachers. In this paper, we propose a new multi-teacher KD-based training strategy, which can incorporate the unbiased fusion knowledge of multi-teachers for robust SGG training under noisy labels.

\textbf{OOD Benchmarks for Visual Scene Understanding.} Recently, several scene understanding tasks, such as VQA~\cite{niu2021introspective,chen2020counterfactual,chen2021counterfactual,chen2022rethinking} and video grounding~\cite{yuan2021closer,lan2022closer}, try to deliberately make the ground-truth distributions different in the training and test splits. For example, Agrawal~\etal~\cite{agrawal2018don} reorganized splits of previous VQA datasets by designing different answer distributions for each question type in the training and test sets. Similarly, another video grounding dataset for OOD evaluation was proposed in a similar manner~\cite{yuan2021closer}. Inspired by them, we \shr{introduce VG-OOD to better measure the OOD generalization abilities of SGG models.}

\section{NICE: NoIsy label CorrEction}

Given an image dataset $\bm{\mathcal{I}}$, the SGG task aims to convert each image $\mathcal{I}_i \in \bm{\mathcal{I}}$ into a graph $\mathcal{G}_i = \{ \mathcal{N}_i,  \mathcal{E}_i \}$, where $\mathcal{N}_i$ and $\mathcal{E}_i$ denote the node set (\ie, objects) and edge set (\ie, visual relations) of image $\mathcal{I}_i$, respectively. Generally, each graph $\mathcal{G}_i$ can also be \shr{viewed} as a set of visual relation triplets (\ie, $\langle$\texttt{subject}-\texttt{predicate}-\texttt{object}$\rangle$), denoted as $\mathcal{T}_i$. For each triplet set $\mathcal{T}_i$, we can further divide it into two subsets: $\mathcal{T}^+_i$ and $\mathcal{T}^-_i$, where $\mathcal{T}^+_i$ denotes all the annotated positive triplets (or samples) in image $\mathcal{I}_i$, and $\mathcal{T}^-_i$ denotes all the un-annotated negative triplets in image $\mathcal{I}_i$. Similarly, we use $\bm{\mathcal{T}}^+ = \{\mathcal{T}^+_i\}$ and $\bm{\mathcal{T}}^- = \{\mathcal{T}^-_i\}$ to represent all positive and negative triplets in the whole dataset $\bm{\mathcal{I}}$.

Fig.~\ref{fig:framework} illustrates the overall pipeline of NICE\footnote{In Fig.~\ref{fig:framework}, we use a single image as input for a clear illustration. In our experiments, we directly process \emph{the whole dataset} in each module. \label{footnote:fig2}}. In this section, we successively introduce each part of NICE, including negative noisy sample detection (Neg-NSD), positive NSD (Pos-NSD), and noisy sample correction (NSC). To be specific, given an image and its corresponding ground-truth triplet annotations (\ie, $\mathcal{T}^+ \bigcup \mathcal{T}^-$)\footnote{For brevity, we omit the subscripts $i$ for image $\mathcal{I}_i$.}, we first use the Neg-NSD to detect all possible noisy negative samples, \ie, missing annotated foreground triplets. $\mathcal{T}^-$ can be further divided into $\mathcal{T}^-_{\text{clean}}$ and $\mathcal{T}^-_{\text{noisy}}$. Meanwhile, Neg-NSD assigns pseudo positive predicate labels for all samples in $\mathcal{T}^-_{\text{noisy}}$ (\eg, \texttt{painted on} for $\langle$\texttt{letter}-\texttt{window}$\rangle$ in Fig.~\ref{fig:framework}). $\mathcal{T}^-_{\text{noisy}}$ with pseudo positive labels and original $\mathcal{T}^+$ compose a new positive set $\widetilde {\mathcal{T}}^+$. Next, we use Pos-NSD to detect all possible noisy positive samples in $\widetilde {\mathcal{T}}^+$, \ie, the positive samples which suffer from either common-prone or synonym-random characteristics (\eg, \texttt{at} for $\langle$\texttt{women}-\texttt{laptop}$\rangle$ in Fig.~\ref{fig:framework}). Similarly, $\widetilde{\mathcal{T}}^+$ can be divided into  $\widetilde{\mathcal{T}}^+_{\text{clean}}$ and $\widetilde{\mathcal{T}}^+_{\text{noisy}}$. Then, we use NSC to reassign more high-quality soft predicate labels to all samples in $\widetilde{\mathcal{T}}^+_{\text{noisy}}$, denoted as $\widetilde{\mathcal{T}}^+_{\text{noisy} \to \text{clean}}$. Lastly, after processing the ground-truth triplet annotations of all images, we can obtain a cleaner version of the dataset ($\widetilde{\bm{\mathcal{T}}}^+_{\text{clean}} \bigcup \widetilde{\bm{\mathcal{T}}}^+_{\text{noisy} \to \text{clean}} \bigcup \bm{\mathcal{T}}^-_{\text{clean}}$) for SGG training.

\subsection{Negative Noisy Sample Detection (Neg-NSD)}

In this module, our goal is to discover all possible \emph{noisy} negative samples, \ie, missing annotated visual relation triplets. Meanwhile, Neg-NSD assigns a pseudo positive predicate label for each detected noisy negative sample. It is difficult to directly train and evaluate a binary classifier based on these noisy samples, due to the nature of missing annotations in existing negative samples. To this end, we propose formulating the negative noisy sample detection as an out-of-distribution (OOD) detection problem~\cite{hendrycks2016baseline}. Specifically, we regard all annotated positive samples as in-distribution (ID) training data, and all unannotated negative samples as OOD test data. Neg-NSD is constructed on top of a plain SGG model (denoted as $\mathtt{F}_{\text{sgg}}^n$), but it is trained exclusively with the annotated positive samples $\bm{\mathcal{T}}^+$. During the inference stage, Neg-NSD \shr{predicts a score of being foreground and assigns} a pseudo positive predicate category to each triplet $t^-_i \in \bm{\mathcal{T}}^-$.

Following existing OOD detection methods~\cite{devries2018learning}, we also utilize a confidence-based model, \ie, Neg-NSD consists of two network output branches: 1) a classification branch to predict a probability distribution $\bm{p}$ over all positive predicate categories, and 2) a confidence branch to predict a confidence score $c \in [0, 1]$, which indicates the confidence of being an ID category (foreground). During the inference phase, for each sample $t^-_i$, if its confidence score $c_i$ surpasses a threshold $\theta$, we then regard this negative sample as a noisy negative sample, \ie, the detection function $g(\cdot)$ is \shr{defined as follows}:
\begin{equation} \label{eq:1}
g(t^-_i) = \left\{ 
\begin{array}{l}
1, \quad c_i \ge \theta \\
0, \quad c_i < \theta.
\end{array} \right.
\end{equation}
When $g(t^-_i) = 1$, the pseudo label of $t^-_i$ is directly derived from the classification branch, \ie, $\arg \max (\bm{p}_i)$. \shr{Due to significant variations in the predicted average confidence scores for different predicate categories}, we set different thresholds for head, body, and tail categories. (More details are in Sec.~\ref{sec:6}.)

\begin{figure}[!t]
  \centering
  \includegraphics[width=\linewidth]{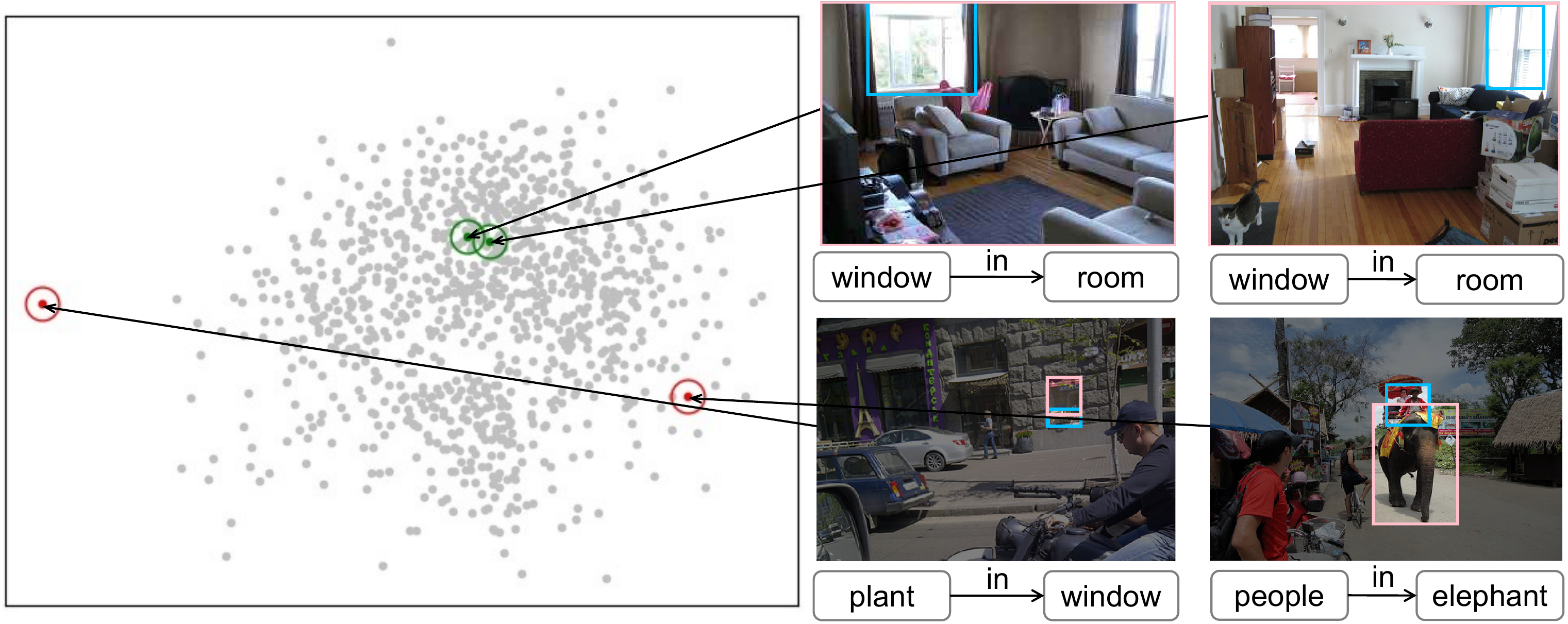}
  \vspace{-1.5em}
  \caption{\textbf{Left}: Multidimensional scaling visualization of features of randomly sampled triplets with predicate \texttt{in}. \textbf{Right}: Detected \textcolor{mygreen}{\textbf{clean}} samples and \textcolor{red}{\textbf{noisy}} samples by Pos-NSD.} 
  \label{fig:example}
  \vspace{-1em}
\end{figure}

\textbf{Training of Neg-NSD.} To train the classification branch and confidence branch, we combine predicted probabilities $\bm{p}_i$ and corresponding target probability distribution $\bm{y}_i$, \ie,
\begin{equation} \label{eq:2}
\bm{p'}_i = c_i \cdot {\bm{p}_i} + (1 - c_i) \cdot \bm{y}_i,
\end{equation}
where $\bm{p'}_i$ is the adjusted probabilities by confidence $c_i$. The motivation of Eq.~\eqref{eq:2} is that\shr{, given the opportunity to request a hint of the ground-truth probability with some penalty,} the model will definitely choose to seek the hint if it \shr{lacks confidence in} its output (\ie, $c_i$ is small). And the training objective for Neg-NSD consists of a weighted cross-entropy loss and a regularization penalty loss: 
\begin{equation}
    \mathcal{L}_{N-NSD} = - \textstyle{\sum}_{j = 1} {{w_{j}}\log ({p'}_{ij}){{y}_{ij}} - \lambda \log (c_{i})}, 
\end{equation}
where $p'_{ij}$ and $y_{ij}$ are the $j$-th elements of $\bm{p'}_i$ and $\bm{y}_i$, respectively. $w_j$ is the reciprocal of the frequency of the $j$-th predicate category, which mitigates the impact of the long-tailed issues on confidence. The penalty loss is used to prevent the network from always choosing $c=0$ and using a ground-truth probability distribution to minimize the task loss.

\subsection{Positive Noisy Sample Detection (Pos-NSD)} \label{sec:3.2}

As shown in Fig.~\ref{fig:framework}\footref{footnote:fig2}, the original positive set $\bm{\mathcal{T}}^+$ and outputs of the Neg-NSD (\ie, $\bm{\mathcal{T}}^-_{\text{noisy}}$) compose a new positive sample set ${\widetilde {\bm{\mathcal{T}}}^+}$. The Pos-NSD module aims to detect all \emph{noisy} samples in ${\widetilde{\bm{\mathcal{T}}}^+}$. In general, we utilize a clustering-based solution to divide all these positive samples into multiple subsets with different degrees of noise, and treat all samples in the noisiest subset as noisy positive samples. Intuitively, if a predicate label is consistent with other visually-similar samples of the same predicate category (\ie, visual features of these samples are close to each other). In that case, this predicate is more likely to be a clean sample since these annotations are consistent with each other. Otherwise, it is likely to be a noisy sample. As shown in Fig.~\ref{fig:example}, the two clean triplets $\langle$\texttt{window}-\texttt{in}-\texttt{room}$\rangle$ have more visually-similar neighbors than the noisy triplets (\eg, $\langle$\texttt{plant}-\texttt{in}-\texttt{window}$\rangle$).

\begin{figure}[!t]
  \centering
  \includegraphics[width=\linewidth]{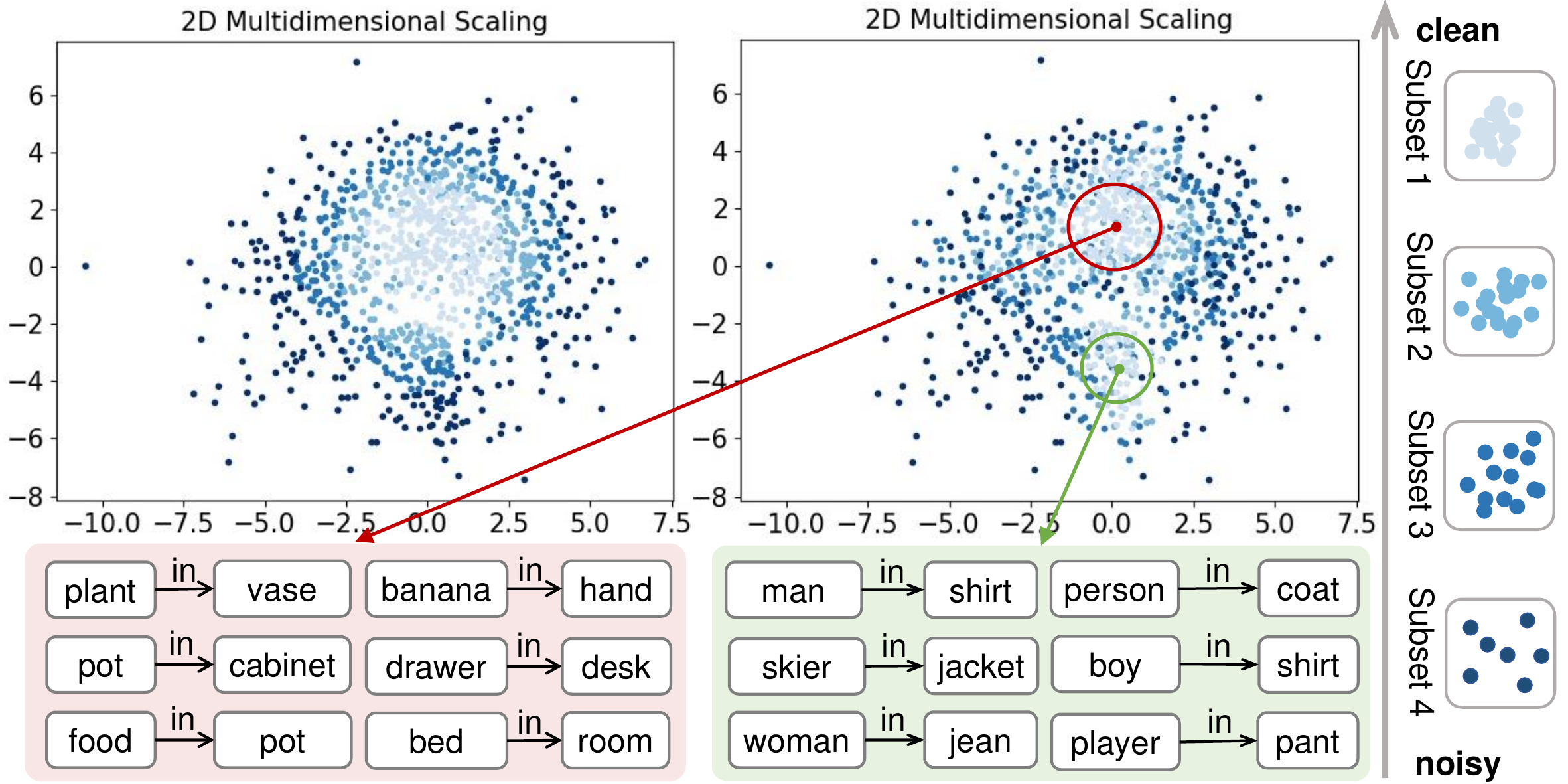}
  \vspace{-1.5em}
  \caption{\textbf{Above}: The multidimensional scaling visualization of features of randomly sampled triplets with predicate \texttt{in} with cutoff distance ranked at 50\% (left) and 1\% (right). \textbf{Below}: The triplet categories of the randomly sampled visual relation triplets from the corresponding \textcolor{red}{\textbf{red}} circle and \textcolor{mygreen}{\textbf{green}} circle.}
  \label{fig:subset}
  \vspace{-1em}
\end{figure}

Based on these observations, we propose a local density based solution for positive noisy sample detection. To be specific, we utilize an off-the-shelf pretrained SGG model (denoted as $\mathtt{F}_{\text{sgg}}^p$) to extract all visual relation triplet features, and utilize $\bm{h}^k_i$ to represent the visual feature of the $i$-th sample of predicate category $k$ (this sample is denoted as $t^k_i$). Then, we use a distance matrix $\bm{D}^k = (d^k_{ij})_{N \times N} \in \mathbb{R}^{N \times N}$ to measure the similarity between all positive samples of the same predicate $k$, and $d^k_{ij}$ is calculated by:
\begin{equation} \label{eq:4}
    d^k_{ij} = \left\| \bm{h}^k_i - \bm{h}^k_j \right\|^2,
\end{equation}
where $\|\cdot\|$ is the Euclidean distance. Thus, a smaller distance $d^k_{ij}$ means a relatively higher similarity between sample $t^k_i$ and sample $t^k_j$. Then following~\cite{rodriguez2014clustering}, we define the local density $\rho^k_i$ of each sample $t^k_i$ as the number of samples (within the same predicate category) whose distances to sample $t^k_i$ are closer than a threshold $d^k_c$, \ie, 
\begin{equation} \label{eq:5}
{\rho^k _i} = \textstyle\sum_j {\mathbf{1} (({d^k_c} - {d^k_{ij}})>0)},
\end{equation}
where $\mathbf{1}(\cdot)$ is the indicator function
and $d^k_c$ is the cutoff distance for predicate $k$, which is ranked at $\alpha\%$ of sorted $N \times N$ distances in $\bm{D}^K$ from small to large. Thus, a sample with higher local density $\rho$ means that this sample is more similar to the samples of the same predicate category. Analogously, samples with low local density $\rho$ are considered as noisy samples. Finally, we use an unsupervised K-means algorithm~\cite{hechenbichler2004weighted} to divide all data samples into multiple subsets with respect to different $\rho$ values, \ie, different degrees of noise~\cite{guo2018curriculumnet}. All samples in the subset with the lowest $\rho$ are regarded as noisy positive samples (\ie, $\widetilde{\bm{\mathcal{T}}}^+_{\text{noisy}}$), and \shr{passed into the subsequent} NSC module for label correction.

\textbf{Influence of the Cutoff Distance \bm{$d^k_c$}.} From Eq.~\eqref{eq:5}, we can observe that the distribution of local density $\rho$ is directly decided by the selection of the cutoff distance $d_c$ (or the hyperparameter $\alpha\%$). As shown in Fig.~\ref{fig:subset}, when the cutoff distance \shr{is} ranked at 50\% and 1\%, local densities of samples diffuse outwards from large to small with one and two centers, respectively, \ie, a smaller cutoff distance (\eg, 1\% for $\alpha\%$) may divide the whole feature space into more clusters. Meanwhile, different predicate categories may contain various types of semantic meanings. For example, in Fig.~\ref{fig:subset}, \shr{the} predicate \texttt{in} of samples inside the red circle represents ``inside" (\eg, $\langle$\texttt{plant}-\texttt{in}-\texttt{vase}$\rangle$), while predicate \texttt{in} of the samples inside the green circle represents ``wearing" (\eg, $\langle$\texttt{man}-\texttt{in}-\texttt{shirt}$\rangle$). Therefore, we set different cutoff distances for different categories. 

In addition, more detailed discussions about the influence of $d^k_c$ on the clustering results are left in the appendix.

\begin{figure}[!t]
  \centering
  \includegraphics[width=\linewidth]{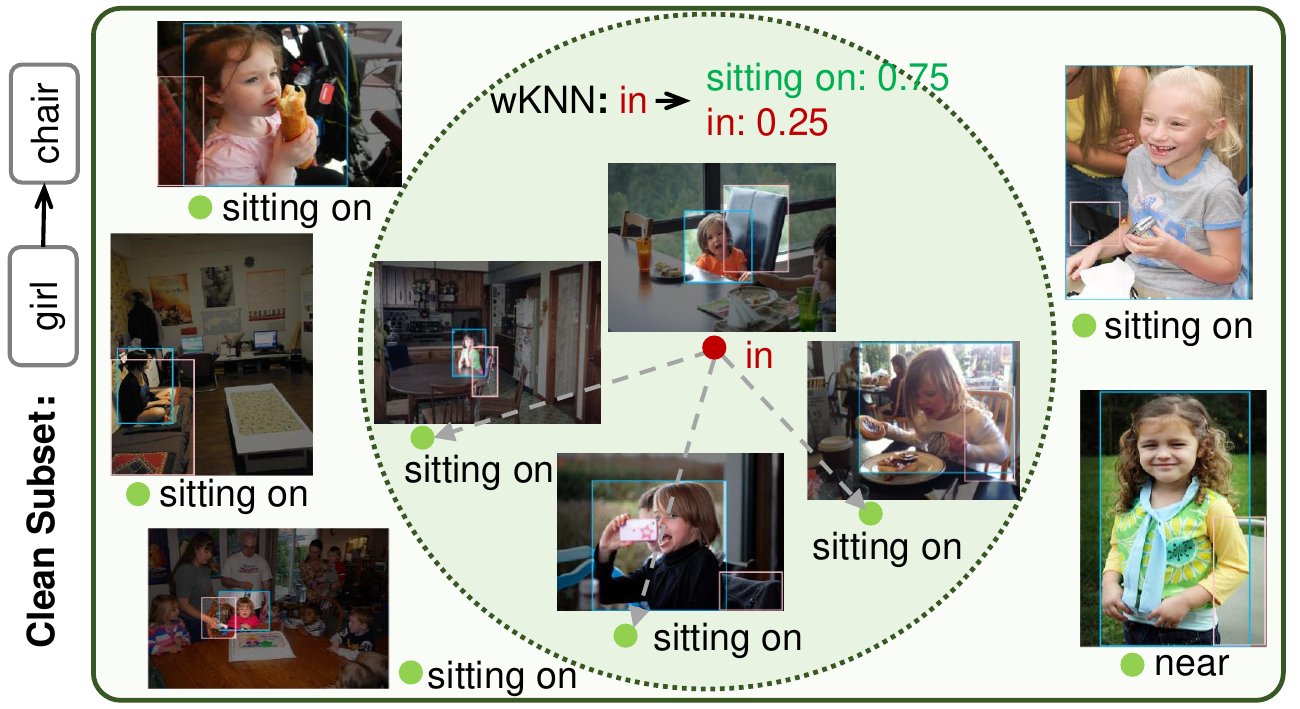}
  \vspace{-1.5em}
   \caption{The illustration of NSC. Dashed lines indicate the distances between the noisy sample and other samples in a clean subset with \texttt{girl}-\texttt{chair}. wKNN replaces the noisy predicate \texttt{in} with a soft label, assigning a score of 0.25 to the new label \texttt{sitting on} and a score of 0.75 to the old label \texttt{in}.}
  \label{fig:correction}
  \vspace{-1em}
\end{figure}

\subsection{Noisy Sample Correction (NSC)}

Given all the detected noisy positive samples from Pos-NSD, the NSC module aims to correct and assign more robust soft labels to these noisy positive predicate labels. There are two non-zero (positive) categories in the newly assigned soft label $r^s$: the original predicate category and a raw predicate category. The raw predicates serve as relatively clean labels assigned by wKNN, \shr{following the approach} in NICE-v1~\cite{li2022devil}. The \shr{subsequent sections elaborate on} the raw predicate categories and the calculation of ground-truth values for the two predicates in the soft labels.

\textbf{Selection of Raw Predicates.} The motivation of our NSC is that the predicate label of a sample should be consistent with other visually-similar training samples, especially for those samples with the same \texttt{subject} and \texttt{object} categories. For example, in Fig.~\ref{fig:correction}, consider the noisy sample $\langle$\texttt{girl}-\texttt{in}-\texttt{chair}$\rangle$. \shr{By retrieving all other samples with the same $\langle$\texttt{girl}-\texttt{chair}$\rangle$, we observe that most visually-similar samples are annotated as $\langle$\texttt{girl}-\texttt{sitting on}-\texttt{chair}$\rangle$.} Thus, we employ the simple yet effective weighted K-Nearest Neighbor (wKNN) algorithm to derive the most probable labels for noisy positive samples. wKNN assigns larger weights to the closest samples and smaller weights to those that are far away. Specifically, let $N(i)$ be the set of $K$ neighbors of sample $t_i$, then the raw predicate label $r^{raw}_{i}$ for $t_i$ is:
\begin{equation}
    {r^{raw}_{i}} = \arg {\max _v}\sum\nolimits_{t_j \in N(i)} {{w_{ij}} \cdot \mathbf{1} (v = {r_j})},
\end{equation}
where $v$ is a predicate category, $r_j$ is predicate label of $t_j$, and $\mathbf{1}(\cdot)$ is an indicator function. The weight $w_{ij}$ is assigned to each neighbor, defined as $a \cdot \exp(- \frac{(d_{ij} - b)^2}{2c^2})$. $d_{ij}$ is the Euclidean distance between $\bm{h}_i$ and $\bm{h}_j$ (Eq.~\eqref{eq:4}), and $a$, $b$, $c$ are hyperparameters. 

\textbf{Calculation of Predicate Score.} However, the raw predicate labels are not always ``perfect''. For example, in Fig.~\ref{fig:example_nice}, $\langle$\texttt{boy}-\texttt{near}-\texttt{pizza}$\rangle$ is mistakenly changed to $\langle$\texttt{boy}-\texttt{eating}-\texttt{pizza}$\rangle$. To ensure the robustness of training, we propose a scoring function \shr{that considers the distances of triplets in weighted KNN as the confidence of the labels and fuses} the raw predicate label $r^{raw}$ with the original predicate label $r^{ori}$ to obtain a new soft label. The \shr{scores} of the raw predicate label $s^{raw}$ and the original predicate label $s^{ori}$ in soft labels are calculated as follows:
\begin{equation}
{s^{raw}} = \frac{{\sum {{w^{raw}}\mathbf{1}(v = {r^{raw}})} }}{{\sum {{w^{raw}}\mathbf{1}(v = {r^{raw}})}  + \sum {{w^{ori}}\mathbf{1}(v = {r^{ori}})} }},
\end{equation}
\begin{equation}
{s^{ori}} = \frac{{\sum {{w^{ori}}\mathbf{1}(v = {r^{ori}})} }}{{\sum {{w^{raw}}\mathbf{1}(v = {r^{raw}})}  + \sum {{w^{ori}}\mathbf{1}(v = {r^{ori}})} }},
\end{equation}
where $w^{raw}$ and $w^{ori}$ represent the weights of the raw predicate and the original predicate in KNN, respectively. 

\textbf{Training Objectives.}
We apply the binary cross entropy objective for training with soft labels, which is written as:
\begin{equation}
 \mathcal{L}_{SL} =  - \frac{1}{N}\sum\limits_i^N {{r_i^s}\log ({p_i}) + (1 - {r_i^s})\log (1 - {p_i})},
\end{equation}
where $r_i^{s}$ is the value of the $i$-th category in the soft label, $p_i$ is the prediction probability of the $i$-th category, and $N$ is the total number of predicate categories.

\noindent\underline{\textbf{Highlights.}} In the original NSC of NICE-v1~\cite{li2022devil}, we directly assign hard labels (\ie, raw predicate labels) with full probability to noisy samples. If the newly assigned labels were unreasonable, new noise might be introduced. In the new NSC of this paper, we assign soft labels to noisy samples, which takes into account both the raw predicate labels by wKNN and the original ground-truth labels. It not only increases the error tolerance rate of noisy sample correction, but also is more conducive to robust training. 

\section{NIST: NoIsy Sample Training}

As mentioned above, the labels corrected by NICE~\cite{li2022devil} are not always accurate. Since most of them are corrected from the head to tail predicates, it may lead to a somewhat over-weighting of tail predicates. We propose a novel multi-teacher knowledge distillation strategy for effective model training that \shr{enables the model to} learn from the unbiased trade-off fusion of two teachers. The structures of both the two teacher models and the student model are the same as the backbone (\ie, Motifs~\cite{zellers2018neural} and VCTree~\cite{tang2019learning}). In this way, we generate a relatively smooth target label that mitigates the performance degradation caused by absolute confidence in the noisy ground-truth with full probability by assigning values to other categories.

\begin{figure}[!t]
  \centering
  \includegraphics[width=\linewidth]{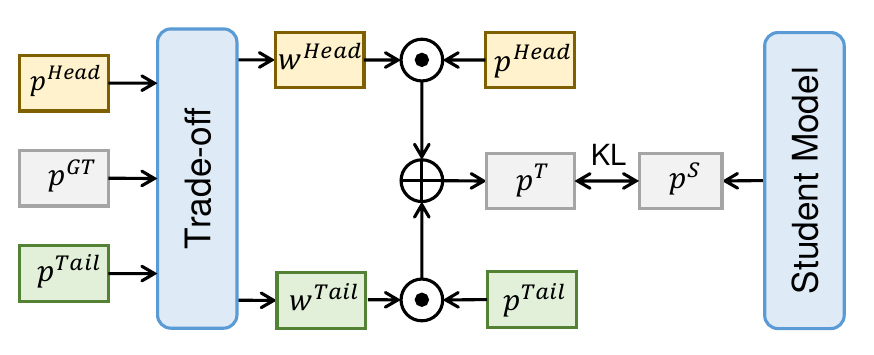}
  \vspace{-2.0em}
  \caption{The pipeline of the multi-teacher KD paradigm.}
  \label{fig:trade_off}
\end{figure}

The pipeline of the multi-teacher knowledge distillation training paradigm is illustrated in Fig.~\ref{fig:trade_off}. Specifically, we initially select two models, \shr{where one excels in} head predicates (\ie, can achieve good performance in Recall) and the other excels in tail predicates (\ie, can achieve good performance in mean Recall). For the sake of simplification, these models are denoted \shr{as} $T^{Head}$ and $T^{Tail}$, respectively. We aim to obtain an unbiased trade-off teacher that can \shr{strike a balance} between Recall and mean Recall (\ie, penalize excessive bias against both head and tail predicates). Therefore, we measure the bias of the model towards head and tail predicates, and define the degrees of biases for head and tail predicates as $s^{Head}$ and $s^{Tail}$, respectively. The calculation process is as follows:
\begin{equation}
{s^{Head} } = \frac{1}{{\mathtt{XE}({p^{GT}},{p^{Head}})}} = \frac{1}{{\sum\nolimits_{j = 1} {{p^{GT}}_j\log {p^{Head}}_j} }},
\end{equation}
\begin{equation}
{s^{Tail} } = \frac{1}{{\mathtt{XE}({p^{\mathop{GT}}},{p^{Tail}})}} = \frac{1}{{\sum\nolimits_{j = 1} {{p^{GT}}_j\log {p^{Tail}}_j} }},
\end{equation}
where $\mathtt{XE}$ denotes the cross-entropy function, and $p^{GT}$, $ p^{Head}$, $ p^{Tail}$ denote the ground-truth labels, the output of $T^{Head}$, and the output of $T^{Tail}$, respectively. We then adopt a dynamic weighting strategy \shr{to fuse the biased knowledge of the two teachers fairly, using} weights $w^{Head}$ and $w^{Tail}$. The final trade-off knowledge $p^{T}$ to supervise the training is determined by:
\begin{equation}
{w^{Head}} = \frac{{{s^{Tail}}}}{{{s^{Head}} + {s^{Tail}}}},
{w^{Tail}} = \frac{{{s^{Head}}}}{{{s^{Head}} + {s^{Tail}}}},
\label{eq:tradeoff}\end{equation}
\begin{equation}
{p^{T}} = {w^{Head}}{p^{Head}} + {w^{Tail}}{p^{Tail}}.
\end{equation}

\textbf{Group Dynamic Weighting.} \shr{Acknowledging that the ground-truth (GT) predicate label is more accurate than $T^{Head}$ for head predicates. Similarly, the GT predicate label is more accurate than $T^{Tail}$ for tail predicates.} Therefore, we \shr{employ a dynamic weighting mechanism for} the head, body, and tail groups to obtain more accurate knowledge for distillation:
\begin{itemize}
    \item For head predicates, $p^{Head}$ is the GT label, and $p^{Tail}$ is the output probability distribution of $p^{Tail}$.
    \item For body predicates, since the GT labels are more accurate than both two models, we directly adopt the GT label as $p^{Head}$ and $p^{Tail}$.
    \item For tail predicates, $p^{Head}$ is the output probability distribution of $p^{Head}$, and $p^{Tail}$ is the GT label.
\end{itemize}

\textbf{Training Objectives.} Since the outputs of the student model and the unbiased teacher model are probability distributions, we adopt KL divergence~\cite{kullback1951information} as the loss function for multi-teacher KD, denoted as:
\begin{equation}
\mathcal{L}_{KD} = \mathtt{KL}(p^{S},p^{T})=\sum\limits_i^N {p_i^{S}\log (\frac{{p^{S}_i}}{{p^{T}_i}})},
\end{equation}
where $p^{S}$ is the output probability distribution of student model.

\noindent\underline{\textbf{Highlights.}} In NICE-v1~\cite{li2022devil}, the original dataset is corrected directly. This correction is not always correct and will exaggerate the proportion of tail predicates. In this paper, we propose a training strategy based on multi-teacher knowledge distillation, which integrates different biased knowledge of teacher models to supervise student model training, allowing the student model to account for the performance of both head and tail predicates.

\begin{figure}[!t]
  \centering
  \includegraphics[width=\linewidth]{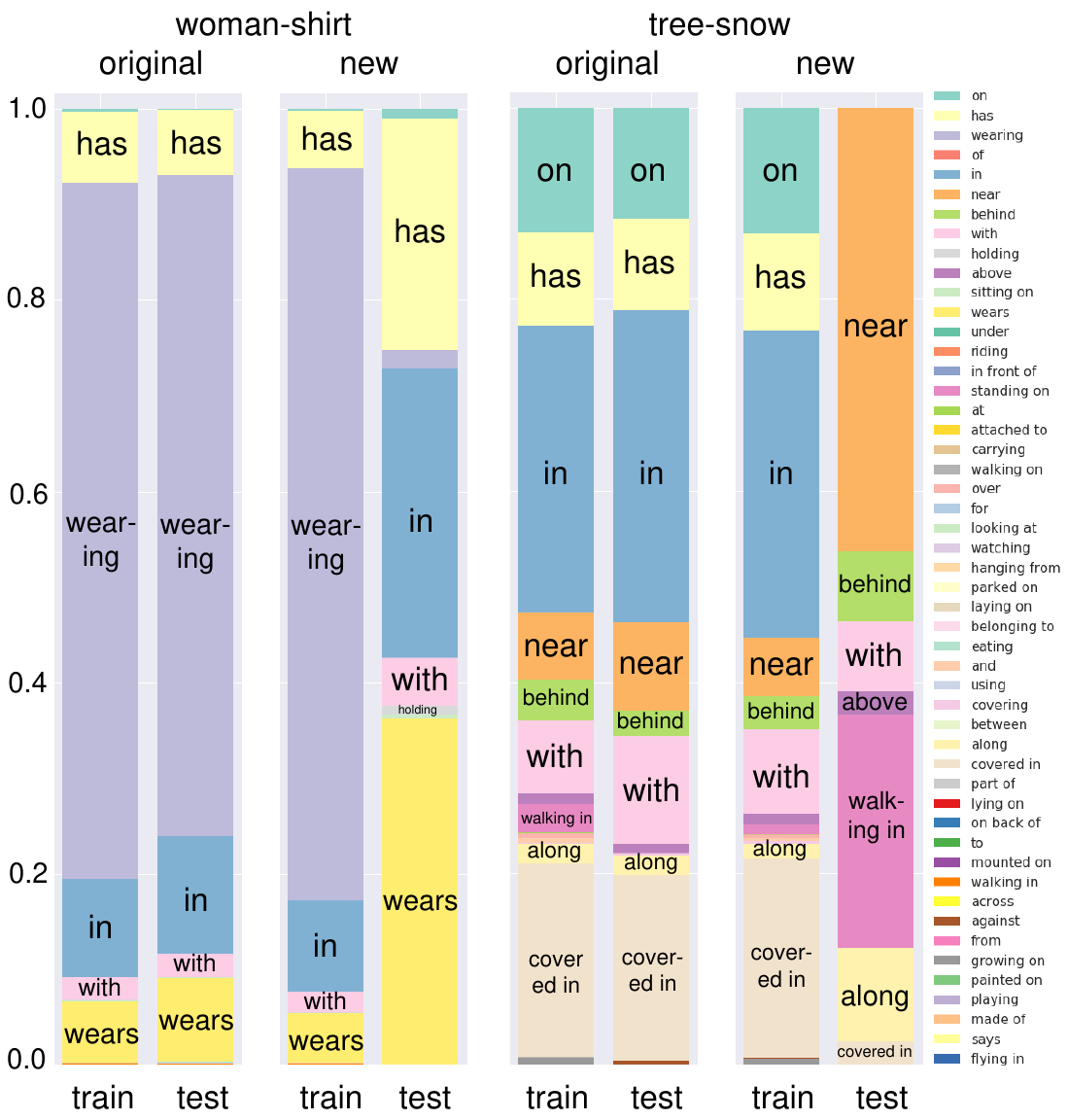}
  \vspace{-1.5em}
  \caption{\textbf{Left}: The predicate statistical distribution of \texttt{woman}-\texttt{shirt}. \textbf{Right}: The predicate statistical distribution of \texttt{tree}-\texttt{snow}. The predicate categories in the legend are sorted by the number of occurrences in the dataset.}
  \label{fig:stat}
\end{figure}

\begin{table}[t]
  \renewcommand\arraystretch{1.2}
  \setlength\tabcolsep{7pt}
  \centering
  \caption{Comparison of VG and VG-OOD datasets. ``KL'' measures KL divergence differences in the predicate distributions of all samples between two splits, and ``KL-mean'' measures the mean of KL divergence differences in the predicate distributions between two splits over all subject-object category pairs.}
  \vspace{-0.5em}
  \resizebox{0.48\textwidth}{!}{
    \renewcommand\arraystretch{1.1}
    \begin{tabular}{c||lrr|c|c}
    \hline\thickhline
    \rowcolor{mygray-bg}
    \multicolumn{1}{l||}{Dataset} & \multicolumn{1}{l|}{Split} & \multicolumn{1}{l|}{\#Image} & \multicolumn{1}{l|}{\#Triplets} & \multicolumn{1}{l|}{KL} & \multicolumn{1}{l}{KL-Mean} \\
    \hline\hline
    \multirow{2}{*}{VG} & Train &  62,723     &  439,063 & \multirow{2}{*}{0.02} & \multirow{2}{*}{0.82} \\
          & Test  & 26,446      &   183,642    &       &  \\
    \hline
    \multirow{2}{*}{VG-OOD} & Train &    64,694   &  521,868  &  \multirow{2}{*}{0.20} & \multirow{2}{*}{7.55} \\
          & Test  &   24,475 &   100,837    &       &  \\
    \specialrule{0.05em}{0pt}{0pt}
    \hline
    \end{tabular}%
    }

  \label{tab:vg_ood}%
\end{table}%

\section{Benchmark: VG-OOD}

Visual Genome~\cite{krishna2017visual}, the most widely utilized SGG dataset at present, has a consistent predicate distribution for each subject-object category pair in both the training and test sets. Some studies~\cite{zellers2018neural} have found that good performance can be obtained just through the frequency prior bias of the commonest predicate category for each subject-object category pair (\ie, 62.2\% on R@100 in Predcls). To reduce the influence of frequency bias and supplement the evaluation of the generalization ability of the model, we propose a new benchmark called VG-OOD by re-splitting the VG dataset. The statistical distributions of all predicates per subject-object category pair (\eg, \texttt{man}-\texttt{shirt}) in VG-OOD are different in the test set compared to the training set.

\textbf{Dataset Construction.}
We first count the number of all possible predicates for each subject-object category pair, and \shr{arrange the predicates based on their frequency in ascending order.} Secondly, we add triplets whose total number is less than 20\% of the \shr{overall} subject-object category pairs in the test triplet list. For example, in Fig.~\ref{fig:stat}, for the \texttt{woman}-\texttt{shirt} pair, we sort the number of predicates and include those with a cumulative number of 20\%, along with their subject-object category pairs, into the list of test triplets (\eg, \texttt{woman}-\texttt{holding}-\texttt{shirt} and \texttt{woman}-\texttt{holding}-\texttt{shirt}). Finally, we split the image into the test set when 70\% of its triplets are in the test triplet list.

\textbf{Dataset Statistics.}
The statistics of the number of images, the number of triplets, and the difference in distribution (\ie, KL and KL-mean) between the VG and VG-OOD training and test sets are shown in TABLE~\ref{tab:vg_ood}. Among them, KL divergence \shr{computes the disparity in predicate distributions between the training and test sets for all samples}. KL-mean is calculated by averaging the KL divergences of the predicate distributions between the two splits across all possible subject-object category pairs. The greater the KL divergence, the more different the two distributions are. Therefore, we can see that \shr{the re-splitting process in VG-OOD results in significant difference in predicate distributions for each subject-object category pair between the training and test sets.} Furthermore, we visualize the distribution of predicates under \texttt{woman}-\texttt{shirt} and \texttt{tree}-\texttt{snow} in Fig.~\ref{fig:stat}. \shr{Notably, there is a marked contrast} in the number of predicates with a large proportion between the \shr{training and test sets} (\eg, \texttt{wearing} \vs 
\texttt{wears}, and \texttt{in} \vs \texttt{near}). \shr{Consequently, relying solely on the frequency statistical prior of triplets does not yield satisfactory performance in the test set.} Moreover, for some triplets only appearing in the test set, the generalization ability of models can be better verified (\eg, \texttt{woman}-\texttt{holding}-\texttt{shirt}). 

\addtolength{\tabcolsep}{-2pt}
\begin{table}[!t]
  \renewcommand\arraystretch{1.2}
  \setlength\tabcolsep{3pt}
  \centering
  \caption{Performance (\%) comparison with and without frequency priors on VG and VG-OOD datasets under the Predcls setting. The baseline model is Motifs~\cite{zellers2018neural}. }
  \vspace{-0.5em}
  \resizebox{0.48\textwidth}{!}{
    \renewcommand\arraystretch{1.0}
    \begin{tabular}{c||c|ccccc}
    \hline\thickhline
    \rowcolor{mygray-bg}
        Dataset & freq prior & mR@50 & mR@100 & R@50  & R@100 & Mean \\
    \hline\hline
        \multirow{2}[2]{*}{VG} 
        & \Checkmark  & 16.5 & 17.8 & 65.5 & 67.2 & 41.8 \\
        & \XSolidBrush &  15.3  & 16.5 &  65.0  & 66.9 & 40.9 \\
    \hline
        \multirow{2}[2]{*}{VG-OOD} 
        & \Checkmark & 11.3  & 11.9  & 57.8  & 58.4  & 34.9 \\
        & \XSolidBrush & 12.5  & 13.0 & 59.8  & 60.5  & 36.5 \\
    \specialrule{0.05em}{0pt}{0pt}
    \hline
    \end{tabular}%
    }
  \label{tab:iid_vs_ood}%
\end{table}%
\addtolength{\tabcolsep}{2pt}

\textbf{VG-OOD vs. VG.}
As mentioned earlier, since the predicate distributions under each subject-object category pair in the training and test sets are consistent in the original VG dataset, a simple frequency bias baseline can achieve decent results. To verify the generalization ability of SGG models to be independent of the frequency bias, we report the performance of models with and without frequency priors in VG and VG-OOD datasets in TABLE~\ref{tab:iid_vs_ood}. It can be observed from TABLE~\ref{tab:iid_vs_ood} that the model can achieve significant gains on VG (\eg, 41.8\% \vs 40.9\% on Mean) by using the frequency prior. However, for the VG-OOD dataset, the model cannot obtain desired results with the frequency prior (\eg, 34.9\% \vs 36.5\% on Mean). Thus, training the model on the VG-OOD dataset can better measure the ability to classify predicates using visual information rather than the frequency bias.

\addtolength{\tabcolsep}{-2pt}
\begin{table*}[tbp]
  \renewcommand\arraystretch{1.2}
  \setlength\tabcolsep{2pt}
    \centering
        \caption{Performance (\%) of state-of-the-art SGG models on three SGG tasks on the VG dataset. ``B" denotes the backbone of object detector (Faster R-CNN~\cite{ren2015faster}) in each SGG model: \ie, VGG-16~\cite{simonyan2014very} and ResNeXt-101-FPN~\cite{lin2017feature}. ``Mean" is the average of mR@50/100 and R@50/100. The \textcolor{red}{\textbf{best}} and \textcolor{blue}{\textbf{second best}} methods under each setting are marked according to formats.}
        \vspace{-1em}
        \resizebox{\textwidth}{!}{
        \renewcommand\arraystretch{1.1}
            \begin{tabular}{c| r ||  c c  c | c  c  c |  c c  c}
                \hline\thickhline
                \rowcolor{mygray-bg}
                &
                 & \multicolumn{3}{c|}{PredCls} & \multicolumn{3}{c|}{SGCls} & \multicolumn{3}{c}{SGGen} \\
                \rowcolor{mygray-bg}
                \multirow{-2}{*}{B} 
                &\multirow{-2}{*}{Models~~~} & \small{mR@50/100} & \small{R@50/100} & \small{Mean} 
               & \small{mR@50/100} & \small{R@50/100} & \small{Mean}  
               & \small{mR@50/100} & \small{R@50/100} & \small{Mean} \\ 
                \hline\hline
               \multirow{4}{*}{\begin{sideways}VGG-16\end{sideways}} 
                & 
                \subt{84}{22}{Motifs~\cite{zellers2018neural}}{ $_{\textit{CVPR'18}}$}
                & 14.0 / 15.3 & 65.2 / 67.1	& 40.4
                & 7.7  / 8.2  & 35.8 / 36.5 & 22.1
                & 5.7  / 6.6  & 27.2 / 30.3 & 17.5 \\
                & \subt{84}{22}{VCTree~\cite{tang2019learning}}{$_{\textit{CVPR'19}}$}
                & 17.9 / 19.4 & 66.4 / 68.1	& 43.0
                & 10.1 / 10.8 & 38.1 / 38.8 & 24.5 
                & 6.9  / 8.0  & 27.9 / 31.3 & 18.5 \\
                & \subt{84}{22}{KERN~\cite{chen2019knowledge}}{$_{\textit{CVPR'19}}$ }
                & 17.7 / 19.2 & 65.8 / 67.6	& 42.6
                & 9.4  / 10.0 & 36.7 / 37.4 & 23.4 
                & 6.4  / 7.3  & 29.8 / 27.1 & 17.7 \\
                & \subt{84}{22}{PCPL~\cite{yan2020pcpl}}{$_{\textit{MM'20}}$ }
                & 35.2 / 37.8 & 50.8 / 52.6	& 44.1
                & 18.6 / 19.6 & 27.6 / 28.4 & 23.6 
                & 9.5  / 11.7 & 14.6 / 18.6 & 13.6 \\
                \hline
                \multirow{18}{*}{\begin{sideways}X-101-FPN\end{sideways}}
                & \subt{84}{22}{MSDN~\cite{li2017scene}}{$_{\textit{ICCV'17}}$}
                & 15.9 / 17.5 & 64.6 / 66.6 & 41.2 
                & 9.3  / 9.7  & 38.4 / 39.8 & 24.3  
                & 6.1  / 7.2  & 31.9 / 36.6 & 20.5 \\
                & \subt{84}{22}{G-RCNN~\cite{yang2018graph}}{$_{\textit{ECCV'18}}$}
                & 16.4 / 17.2 & 64.8 / 66.7 & 41.3 
                & 9.0  / 9.5  & 38.5 / 37.0 & 23.5  
                & 5.8  / 6.6  & 29.7 / 32.8 & 18.7 \\
                & \subt{84}{22}{BGNN~\cite{li2021bipartite}}{$_{\textit{CVPR'21}}$}
                & 30.4 / 32.9 & 59.2 / 61.3 & 45.9 
                & 14.3 / 16.5 & 37.4 / 38.5 & 26.7 
                & 10.7 / 12.6 & 31.0 / 35.8 & 22.5 \\
                & \subt{84}{22}{DT2-ACBS~\cite{desai2021learning}}{$_{\textit{ICCV'21}}$}
                & 35.9 / 39.7 & 23.3 / 25.6 & 31.1
                & 24.8 / 27.5 & 16.2 / 17.6 & 21.5 
                & 22.0 / 24.4 & 15.0 / 16.3 & 19.4 \\ 
                \cline{2-11}
                & \subt{84}{22}{Motifs~~\cite{zellers2018neural}}{$_{\textit{CVPR'18}}$}
                & 16.5 / 17.8 & \textcolor{red}{\textbf{65.5}} / \textcolor{red}{\textbf{67.2}} & 41.8
                & 8.7  / 9.3  & \textcolor{red}{\textbf{39.0}} / \textcolor{red}{\textbf{39.7}} & 24.2
                & 5.5  / 6.8  & \textcolor{red}{\textbf{32.1}} / \textcolor{red}{\textbf{36.9}} & 20.3 \\ 
                &  \subt{84}{22}{Motifs+TDE~\cite{tang2020unbiased}}{$_{\textit{CVPR'20}}$}
                & 24.2 / 27.9 & 45.0 / 50.6 & 36.9
                & 13.1 / 14.9 & 27.1 / 29.5 & 21.2 
                & 9.2  / 11.1 & 17.3 / 20.8 & 14.6 \\
                & \subt{84}{22}{Motifs+PCPL~\cite{yan2020pcpl}}{$_{\textit{MM'20}}$}
                & 24.3 / 26.1 & 54.7 / 56.5 & 40.4
                & 12.0 / 12.7 & \textcolor{blue}{\textbf{35.3}} / \textcolor{blue}{\textbf{36.1}} & 24.0 
                & 10.7 / 12.6 & 27.8 / 31.7 & 20.7 \\
                & \subt{84}{22}{Motifs+CogTree~\cite{yu2021cogtree}}{$_{\textit{IJCAI'21}}$}
                & 26.4 / 29.0 & 35.6 / 36.8	& 32.0 
                & 14.9 / 16.1 & 21.6 / 22.2 & 18.7 	
                & 10.4 / 11.8 & 20.0 / 22.1 & 16.1 \\
                & \subt{84}{22}{Motifs+DLFE~\cite{chiou2021recovering}}{$_{\textit{MM'21}}$ }
                & 26.9 / 28.8 & 52.5 / 54.2 & 40.6
                & 15.2 / 15.9 & 32.3 / 33.1 & 24.1
                & 11.7 / 13.8 & 25.4 / 29.4 & 20.1 \\
                & \subt{84}{22}{Motifs+BPL-SA~\cite{guo2021general}}{$_{\textit{ICCV'21}}$}
                & 29.7 / 31.7 & 50.7 / 52.5 & 41.2
                & \textcolor{blue}{\textbf{16.5}} / 17.5 & 30.1 / 31.0 & 23.8
                & \textcolor{red}{\textbf{13.5}} / \textcolor{red}{\textbf{15.6}} & 23.0 / 26.9 & 19.8 \\ 
                & \cellcolor{mygray-bg}{\subt{84}{22}{Motifs+NICE-v1~\cite{li2022devil}}{$_{\textit{CVPR'22}}$}}
                & \cellcolor{mygray-bg}{\textcolor{blue}{\textbf{29.9}}} / \cellcolor{mygray-bg}{\textcolor{red}{\textbf{32.3}}} & \cellcolor{mygray-bg}{55.1} / \cellcolor{mygray-bg}{57.2} & \cellcolor{mygray-bg}{43.6}
                & \cellcolor{mygray-bg}{\textcolor{red}{\textbf{16.6}}} / \cellcolor{mygray-bg}{\textcolor{red}{\textbf{17.9}}} & \cellcolor{mygray-bg}{33.1} / \cellcolor{mygray-bg}{34.0} & \cellcolor{mygray-bg}{25.4}
                & \cellcolor{mygray-bg}{\textcolor{blue}{\textbf{12.2}}} / \cellcolor{mygray-bg}{\textcolor{blue}{\textbf{14.4}}} & \cellcolor{mygray-bg}{27.8} / \cellcolor{mygray-bg}{31.8} & \cellcolor{mygray-bg}{\textcolor{red}{\textbf{21.6}}} \\
                & \cellcolor{mygray-bg}{\textbf{Motifs+NICE (Ours)}} 
                & \cellcolor{mygray-bg}{\textcolor{red}{\textbf{30.0}}} / \cellcolor{mygray-bg}{\textcolor{blue}{\textbf{32.1}}} &  \cellcolor{mygray-bg}{56.6} / \cellcolor{mygray-bg}{58.6} &  \cellcolor{mygray-bg}{\textcolor{blue}{\textbf{44.3}}}
                &  \cellcolor{mygray-bg}{16.4} /  \cellcolor{mygray-bg}{\textcolor{blue}{\textbf{17.5}}} &  \cellcolor{mygray-bg}{33.8} / \cellcolor{mygray-bg}{34.7} &  \cellcolor{mygray-bg}{\textcolor{red}{\textbf{25.6}}}
                &  \cellcolor{mygray-bg}{10.4} /  \cellcolor{mygray-bg}{12.7} &  \cellcolor{mygray-bg}{27.8} / \cellcolor{mygray-bg}{32.0} &  \cellcolor{mygray-bg}{20.7} \\
                & \cellcolor{mygray-bg}{\textbf{Motifs+NICEST (Ours)}} 
                &  \cellcolor{mygray-bg}{29.5} / \cellcolor{mygray-bg}{31.6} &  \cellcolor{mygray-bg}\textcolor{blue}{\textbf{59.1}} / \cellcolor{mygray-bg}\textcolor{blue}{\textbf{61.0}} &  \cellcolor{mygray-bg}{\textcolor{red}{\textbf{45.3}}}
                &  \cellcolor{mygray-bg}{15.7} /  \cellcolor{mygray-bg}{16.5} &  \cellcolor{mygray-bg}{34.4} / \cellcolor{mygray-bg}{35.2} &  \cellcolor{mygray-bg}{\textcolor{blue}{\textbf{25.5}}}
                &  \cellcolor{mygray-bg}{10.4} /  \cellcolor{mygray-bg}{12.4} &  \cellcolor{mygray-bg}\textcolor{blue}{\textbf{28.0}} / \cellcolor{mygray-bg}\textcolor{blue}{\textbf{32.4}} &  \cellcolor{mygray-bg}{\textcolor{blue}{\textbf{20.8}}} \\
                \cline{2-11}
                & \subt{84}{22}{VCTree~\cite{tang2019learning}}{$_{\textit{CVPR'19}}$}
                & 17.1 / 18.4 & \textcolor{red}{\textbf{65.9}} / \textcolor{red}{\textbf{67.5}} & 42.2
                & 10.8 / 11.5 & \textcolor{red}{\textbf{45.6}} / \textcolor{red}{\textbf{46.5}} & 28.6 
                & 7.2  / 8.4  & \textcolor{red}{\textbf{32.0}} / \textcolor{red}{\textbf{36.2}} & 20.9 \\
                & \subt{84}{22}{VCTree+TDE~\cite{tang2020unbiased}}{$_{\textit{CVPR'20}}$}
                & 26.2 / 29.6 & 44.8 / 49.2 & 37.5
                & 15.2 / 17.5 & 28.8 / 32.0 & 23.4 
                & 9.5  / 11.4 & 17.3 / 20.9 & 14.8 \\
                & \subt{84}{22}{VCTree+PCPL~\cite{yan2020pcpl}}{$_{\textit{MM'20}}$}
                & 22.8 / 24.5 & 56.9 / 58.7 & 40.7
                & 15.2 / 16.1 & 40.6 / 41.7 & 28.4  
                & 10.8 / 12.6 & 26.6 / 30.3 & 20.1 \\
                & \subt{84}{22}{VCTree+CogTree~\cite{yu2021cogtree}}{$_{\textit{IJCAI'21}}$ }
                & 27.6 / 29.7 & 44.0 / 45.4 & 36.7 
                & 18.8 / 19.9 & 30.9 / 31.7 & 25.3 
                & 10.4 / 12.1 & 18.2 / 20.4 & 15.3 \\
                & \subt{84}{22}{VCTree+DLFE~\cite{chiou2021recovering}}{$_{\textit{MM'21}}$}
                & 25.3 / 27.1 & 51.8 / 53.5 & 39.4 
                & 18.9 / 20.0 & 33.5 / 34.6 & 26.8 
                & 11.8 / 13.8 & 22.7 / 26.3 & 18.7 \\ 
                & \subt{84}{22}{VCTree+BPL-SA~\cite{guo2021general}}{$_{\textit{ICCV'21}}$}
                & 30.6 / 32.6 & 50.0 / 51.8 & 41.3
                & \textcolor{red}{\textbf{20.1}} / 21.2 & 34.0 / 35.0 & 27.6
                & \textcolor{red}{\textbf{13.5}} / \textcolor{red}{\textbf{15.7}} & 21.7 / 25.5 & 19.1 \\ 
                & \cellcolor{mygray-bg}{\subt{84}{22}{VCTree+NICE-v1~\cite{li2022devil}}{$_{\textit{CVPR'22}}$}}
                & \cellcolor{mygray-bg}{\textcolor{blue}{\textbf{30.7}}} / \cellcolor{mygray-bg}{\textcolor{blue}{\textbf{33.0}}} & \cellcolor{mygray-bg}{55.0} / \cellcolor{mygray-bg}{56.9} & \cellcolor{mygray-bg}{43.9} 
                & \cellcolor{mygray-bg}{19.9} / \cellcolor{mygray-bg}{\textcolor{red}{\textbf{21.3}}} & \cellcolor{mygray-bg}{37.8} / \cellcolor{mygray-bg}{39.0} & \cellcolor{mygray-bg}{\textcolor{blue}{\textbf{29.5}}}
                & \cellcolor{mygray-bg}{\textcolor{blue}{\textbf{11.9}}} / \cellcolor{mygray-bg}{\textcolor{blue}{\textbf{14.1}}} & \cellcolor{mygray-bg}{27.0} / \cellcolor{mygray-bg}{30.8} & \cellcolor{mygray-bg}{\textcolor{blue}{\textbf{21.0}}} \\
                & \cellcolor{mygray-bg}{\textbf{VCTree+NICE (Ours)}} 
                & \cellcolor{mygray-bg}{\textcolor{red}{\textbf{30.9}}} / \cellcolor{mygray-bg}{\textcolor{red}{\textbf{33.1}}} &  \cellcolor{mygray-bg}{56.4} / \cellcolor{mygray-bg}{58.3} & \cellcolor{mygray-bg}{{\textcolor{blue}{\textbf{44.7}}}}& \cellcolor{mygray-bg}\textcolor{blue}{\textbf{20.0}} / \textcolor{blue}{\textbf{21.2}}& \cellcolor{mygray-bg}\textcolor{blue}{\textbf{38.7}} / \cellcolor{mygray-bg}\textcolor{blue}{\textbf{39.8}} & 
                \cellcolor{mygray-bg}\textcolor{red}{\textbf{29.9}}	& \cellcolor{mygray-bg}{10.1} / \cellcolor{mygray-bg}{12.1}	&  \cellcolor{mygray-bg}{28.4} / \cellcolor{mygray-bg}{32.6} & \cellcolor{mygray-bg}{20.8} \\
                & \cellcolor{mygray-bg}{\textbf{VCTree+NICEST (Ours)}} 
                & \cellcolor{mygray-bg}{30.6} / \cellcolor{mygray-bg}{32.9} &  \cellcolor{mygray-bg}\textcolor{blue}{\textbf{59.1}} / \cellcolor{mygray-bg}\textcolor{blue}{\textbf{60.9}} & \cellcolor{mygray-bg}{{\textcolor{red}{\textbf{45.9}}}}& \cellcolor{mygray-bg}{18.9} / \cellcolor{mygray-bg}{20.0} & \cellcolor{mygray-bg}{38.4} / \cellcolor{mygray-bg}{39.4} & \cellcolor{mygray-bg}{29.2} & \cellcolor{mygray-bg}{10.2} / \cellcolor{mygray-bg}{11.9}	&  \cellcolor{mygray-bg}{\textcolor{blue}{\textbf{29.0}}} / \cellcolor{mygray-bg}{\textcolor{blue}{\textbf{32.7}}} & \cellcolor{mygray-bg}{\textcolor{red}{\textbf{\textbf{21.0}}}} \\
                \specialrule{0.05em}{0pt}{0pt}
                \hline
            \end{tabular}}

    \label{tab:compare_with_sota}
\end{table*}
\addtolength{\tabcolsep}{2pt}
\addtolength{\tabcolsep}{-2pt}
\begin{table*}[t]
  \renewcommand\arraystretch{1.2}
  \setlength\tabcolsep{5pt}
    \centering
    \caption{Performance (\%) of state-of-the-art SGG models on three SGG tasks on the VG-OOD dataset. ``Mean" is the average of mR@50/100 and R@50/100. The \textcolor{red}{\textbf{best}} and \textcolor{blue}{\textbf{second best}} methods under each setting are marked according to formats.}
    \label{tab:compare_with_sota_new_split}
    \vspace{-1em}
        \resizebox{\textwidth}{!}{
        \renewcommand\arraystretch{1.1}
            \begin{tabular}{ r||ccc|ccc|ccc}
                \hline\thickhline
                \rowcolor{mygray-bg}
                & \multicolumn{3}{c|}{PredCls} & \multicolumn{3}{c|}{SGCls} & \multicolumn{3}{c}{SGGen}\\
                \rowcolor{mygray-bg}
                \multirow{-2}{*}{Models~~~} & {mR@50/100} & {R@50/100} & {Mean} 
               & {mR@50/100} & {R@50/100} & {Mean}  
               & {mR@50/100} & {R@50/100} & {Mean} \\ 
                \hline \hline 
                \subt{84}{22}{Motifs~\cite{zellers2018neural}}{$_{\textit{CVPR'18}}$}
                & 11.3 / 11.9 & \textcolor{red}{\textbf{57.8}} / \textcolor{red}{\textbf{58.4}} & 34.9
                & 8.0  / 8.5  & \textcolor{red}{\textbf{39.1}} / \textcolor{red}{\textbf{39.9}} & 23.9
                & 5.5  / 6.8  & \textcolor{red}{\textbf{32.1}} / \textcolor{red}{\textbf{36.9}} & 20.3 \\ 
                \subt{84}{22}{Motifs+TDE~\cite{tang2020unbiased}}{{$_{\textit{CVPR'20}}$}}
                & 25.1 / 27.1 & 47.9 / 50.8 & 37.7
                & 12.9 / 13.5 & 32.0 / 33.2 & 22.9 
                & 9.2  / 10.8 & 18.7 / 22.3 & 15.3 \\
                \rowcolor{mygray-bg}
                 \subt{84}{22}{Motifs+NICE-v1 ~\cite{li2022devil}}{{$_{\textit{CVPR'22}}$}}

                &  \cellcolor{mygray-bg}{\textcolor{red}{\textbf{28.4}} / \textcolor{red}{\textbf{29.2}}} &  \cellcolor{mygray-bg}{47.9 / 48.6} &  \cellcolor{mygray-bg}{38.5}
                &  \cellcolor{mygray-bg}{\textcolor{blue}{\textbf{16.9}} / \textcolor{blue}{\textbf{18.1}}} &  \cellcolor{mygray-bg}{32.9 / 33.9} & \cellcolor{mygray-bg}{\textcolor{blue}{\textbf{25.5}}}
                &  \cellcolor{mygray-bg}{\textcolor{red}{\textbf{12.5}} / \textcolor{red}{\textbf{14.6}}} &  \cellcolor{mygray-bg}{27.0 / 30.4} &  \cellcolor{mygray-bg}{\textcolor{blue}{\textbf{21.1}}} \\
                \cellcolor{mygray-bg}{\textbf{Motifs+NICE (Ours)}} 
                &  \cellcolor{mygray-bg}{\textcolor{blue}{\textbf{28.0}}} / \cellcolor{mygray-bg}{\textcolor{blue}{\textbf{28.9}}} &  \cellcolor{mygray-bg}{50.5} / \cellcolor{mygray-bg}{51.3} &  \cellcolor{mygray-bg}{\textcolor{blue}{\textbf{39.7}}}
                &  \cellcolor{mygray-bg}{15.4} /  \cellcolor{mygray-bg}{15.9} &  \cellcolor{mygray-bg}{34.5} / \cellcolor{mygray-bg}{34.9} &  \cellcolor{mygray-bg}{25.2}
                &  \cellcolor{mygray-bg}{9.6} /  \cellcolor{mygray-bg}{11.2} &  \cellcolor{mygray-bg}\textcolor{blue}{\textbf{28.8}} / \cellcolor{mygray-bg}\textcolor{blue}{\textbf{33.0}} &  \cellcolor{mygray-bg}{20.7} \\
                \cellcolor{mygray-bg}{\textbf{Motifs+NICEST (Ours)}}
                & 
                \cellcolor{mygray-bg}{26.4} / \cellcolor{mygray-bg}{27.1} &  \cellcolor{mygray-bg}\textcolor{blue}{\textbf{54.8}} / \cellcolor{mygray-bg}\textcolor{blue}{\textbf{55.5}} &  \cellcolor{mygray-bg}{\textcolor{red}{\textbf{41.0}}}
                &  \cellcolor{mygray-bg}{\textcolor{red}{\textbf{17.0}}} /  \cellcolor{mygray-bg}{\textcolor{red}{\textbf{18.8}}} &  \cellcolor{mygray-bg}{\textcolor{blue}{\textbf{34.7}}} / \cellcolor{mygray-bg}{\textcolor{blue}{\textbf35.0}} &  \cellcolor{mygray-bg}{\textcolor{red}{\textbf{26.4}}}
                &  \cellcolor{mygray-bg}\textcolor{blue}{\textbf{12.2}} /  \cellcolor{mygray-bg}\textcolor{blue}{\textbf{14.4}} &  \cellcolor{mygray-bg}{27.8} / \cellcolor{mygray-bg}{31.8} &  \cellcolor{mygray-bg}{\textcolor{red}{\textbf{21.6}}} \\
                \cline{1-10}
                 VCTree~\cite{tang2019learning}$_{\textit{CVPR'19}}$
                & 12.1 / 12.5 & \textcolor{red}{\textbf{58.2}} / \textcolor{red}{\textbf{58.8}} & 35.4
                & 8.2 / 8.4 & \textcolor{red}{\textbf{43.2}} / \textcolor{red}{\textbf{43.6}} & 25.9 
                & 5.8  / 6.6  & \textcolor{red}{\textbf{31.9}} / \textcolor{red}{\textbf{35.5}} & 20.0 \\
                 VCTree+TDE~\cite{tang2020unbiased}$_{\textit{CVPR'20}}$
                & 24.6 / 26.0 & 48.7 / 50.9 & 37.6
                & 13.3 / 14.7 & 33.4 / 35.1 & 24.1 
                & 9.6  / 11.4 & 19.4 / 23.1 & 15.9 \\

                \cellcolor{mygray-bg}{VCTree+NICE-v1~\cite{li2022devil}$_{\textit{CVPR'22}}$}
                & \cellcolor{mygray-bg}{\textcolor{blue}{\textbf{28.6}}} / \cellcolor{mygray-bg}{\textcolor{blue}{\textbf{29.4}}} & \cellcolor{mygray-bg}{47.7} / \cellcolor{mygray-bg}{48.4} & \cellcolor{mygray-bg}{38.5} 
                & \cellcolor{mygray-bg}{\textcolor{blue}{\textbf{18.8}}} / \cellcolor{mygray-bg}{\textcolor{blue}{\textbf{19.2}}} & \cellcolor{mygray-bg}{35.3} / \cellcolor{mygray-bg}{35.8} & \cellcolor{mygray-bg}{27.3} 
                & \cellcolor{mygray-bg}{\textcolor{red}{\textbf{12.3}}} / \cellcolor{mygray-bg}{\textcolor{red}{\textbf{14.1}}} & \cellcolor{mygray-bg}{26.0} / \cellcolor{mygray-bg}{29.2} & \cellcolor{mygray-bg}{20.4} \\
                
                \cellcolor{mygray-bg}{\textbf{VCTree+NICE (Ours)}} 
                &
                \cellcolor{mygray-bg}{\textcolor{red}{\textbf{28.6}}} / \cellcolor{mygray-bg}{\textcolor{red}{\textbf{29.5}}} &  \cellcolor{mygray-bg}{50.0} / \cellcolor{mygray-bg}{50.8} & \cellcolor{mygray-bg}{{\textcolor{blue}{\textbf{39.7}}}}& \cellcolor{mygray-bg}{17.7} / \cellcolor{mygray-bg}{18.1}& \cellcolor{mygray-bg}{38.2} / \cellcolor{mygray-bg}{38.6} & 
                \cellcolor{mygray-bg}{\textcolor{blue}{\textbf{28.2}}}	& \cellcolor{mygray-bg}{9.6} / \cellcolor{mygray-bg}{11.3}	&  \cellcolor{mygray-bg}{28.2} / \cellcolor{mygray-bg}{32.3} & \cellcolor{mygray-bg}{\textcolor{blue}{\textbf{\textbf{20.4}}}} \\
                
                \cellcolor{mygray-bg}{\textbf{VCTree+NICEST (Ours)}}
                &
                \cellcolor{mygray-bg}{26.6} / \cellcolor{mygray-bg}{27.4} &  \cellcolor{mygray-bg}\textcolor{blue}{\textbf{54.3}} / \cellcolor{mygray-bg}\textcolor{blue}{\textbf{55.0}} & \cellcolor{mygray-bg}{{\textcolor{red}{\textbf{40.8}}}}& \cellcolor{mygray-bg}\textcolor{red}{\textbf{18.9}} / \cellcolor{mygray-bg}{\textcolor{red}{\textbf{20.0}}}& \cellcolor{mygray-bg}\textcolor{blue}{\textbf{38.4}} / \cellcolor{mygray-bg}\textcolor{blue}{\textbf{39.4}} & 
                \cellcolor{mygray-bg}{\textcolor{red}{\textbf{29.2}}}	& \cellcolor{mygray-bg}\textcolor{blue}{\textbf{10.2}} / \cellcolor{mygray-bg}\textcolor{blue}{\textbf{11.8}}	&  \cellcolor{mygray-bg}\textcolor{blue}{\textbf{29.0}} / \cellcolor{mygray-bg}\textcolor{blue}{\textbf{32.7}} & \cellcolor{mygray-bg}{\textcolor{red}{\textbf{\textbf{20.9}}}} \\ 
                \specialrule{0.05em}{0pt}{0pt}
                \hline
            \end{tabular}
        }
\end{table*}
\addtolength{\tabcolsep}{2pt}

\section{Experiments}
\label{sec:6}
\subsection{Experimental Settings and Implementation Details}

\textbf{Datasets.} To ensure a comprehensive evaluation of our proposed method, we conducted all experiments on three datasets: the challenging VG~\cite{krishna2017visual}, our newly split VG-OOD and GQA~\cite{hudson2019gqa}. 1) \textbf{VG}: VG is the most widely utilized benchmark for SGG with over 108k images. We selected VG to thoroughly evaluate NICEST and ensure a fair comparison with SOTA methods. We followed widely-used splits~\cite{xu2017scene}, dividing images into training (70\%), testing (30\%), and a validation set (5,000 images sampled from the training set~\cite{zellers2018neural}). Besides, we followed~\cite{liu2019large} to divide all predicate categories into three parts based on the number of samples in the training set: head ($>$10k), body (0.5k$\sim$10k), and tail ($<$0.5k). 2) \textbf{VG-OOD}: VG-OOD, our newly proposed dataset, has been deliberately structured by re-splitting the original VG dataset to make the predicate distribution of each subject-object category pair in the training and test sets as inconsistent as possible. To verify the ability to generalize independently of frequency bias, we conducted experiments on VG-OOD and adopted the same predicate grouping as VG. 3) \textbf{GQA}: A large-scale SGG dataset, which consists of more than 110k images. To validate the effectiveness and generalizability of NICEST, we conducted experiments on GQA. We used the same split provided by~\cite{dong2022stacked}, which includes 200 object categories and 100 predicate categories.


\textbf{Evaluation Tasks.} We evaluated NICEST on three SGG tasks~\cite{xu2017scene}: 1) \emph{Predicate Classification} (\textbf{PredCls}): Given the ground-truth objects with labels, we need to only predict pairwise predicate categories. 2) \emph{Scene Graph Classification} (\textbf{SGCls}): Given the ground-truth object bounding boxes, we need to predict both the object categories and predicate categories. 3) \emph{Scene Graph Generation} (\textbf{SGGen}): Given an image, we need to detect all object bounding boxes, and predict both the object categories and predicate categories.

\textbf{Evaluation Metrics.} We evaluated all results on three metrics: 1) \emph{Recall@K} (\textbf{R@K}): It calculates the proportion of top-K confident predicted relation triplets that are in the ground-truth. Following prior work, we used $K = \{50, 100\}$. 2) \emph{mean Recall@K} (\textbf{mR@K}): It calculates the recall for each predicate category separately, and then averages R@K over all predicates, \ie, it puts relatively more emphasis on the tail categories. 3) \textbf{Mean}: It is the mean of all mR@K and R@K scores~\cite{li2022devil}. R@K favors head predicates, while mR@K favors tail ones. Therefore, it is a comprehensive metric that can better reflect model performance on different predicates.

\textbf{NICE Training Details.} In Neg-NSD, we used the Motifs~\cite{zellers2018neural} as OOD detection model $\mathtt{F}_{\text{sgg}}^n$. The training settings (\eg, learning rate and batch size) \shr{followed the configurations} of~\cite{tang2020unbiased} under the PredCls task, except that it was trained with only foreground samples. In Pos-NSD, we used a pretrained Motifs~\cite{zellers2018neural} provided by~\cite{tang2020unbiased} as $\mathtt{F}_{\text{sgg}}^p$ to extract triplet features (\cf, $\bm{h}_i^k$ in Eq.~\eqref{eq:4}) under PredCls task. The number of divided subsets was set to 4. In NSC, the $a$, $b$, and $c$ were set to 1, 0, and 10, respectively. Note that though we used two Motifs models (one is an off-the-shelf model~\cite{zellers2018neural, tang2020unbiased}) in NICE, we only needed to train NICE once, and we could use the obtained cleaner annotations for any model.

\textbf{NIST Training Details.} In the experiment, the models with the bias of head predicates were the widely used baseline models (\ie, Motifs~\cite{zellers2018neural} and VCTree~\cite{tang2019learning}). The models with the bias of tail predicates were trained after NICE cleaning (\ie, Moitfs+NICE and VCTree+NICE). The models obtained by NIST strategy training were denoted as Motifs+NICEST and VCtree+NICEST, respectively. With the implementation of multi-teacher knowledge distillation, our proposed method incurs twice the time and GPU space during the training stage compared to the baseline. However, there is no additional overhead during the inference stage.

\textbf{SGG Training Details.} Our NICEST is a model-agnostic strategy, and thus, for different baselines (\eg, Motifs and VCTree), we followed their respective configurations. We adopted the Faster R-CNN~\cite{ren2015faster} with the ResNeXt-101-FPN~\cite{lin2017feature} backbone to detect objects in the image. The parameters of the detector were frozen during training. Besides, we utilized the SGG benchmark provided by~\cite{tang2020unbiased} for all baselines.
\vspace{-5pt}


\subsection{Comparisons with State-of-the-Arts}
\subsubsection{Performance on VG}
\textbf{Settings.} Since NICE and NIST are model-agnostic strategies, they can be seamlessly incorporated into any advanced SGG model. In this section, we equipped NICE and NIST into two baselines: \textbf{Motifs}~\cite{zellers2018neural} and \textbf{VCTree}~\cite{tang2019learning}, and compared them with the state-of-the-art SGG methods. According to the generalization of these methods, we group them into two categories: 1) \textbf{TDE}~\cite{tang2020unbiased}, \textbf{PCPL}~\cite{yan2020pcpl}, \textbf{CogTree}~\cite{yu2021cogtree}, \textbf{DLFE}~\cite{chiou2021recovering}, \textbf{BPL-SA}~\cite{guo2021general} and our previous work \textbf{NICE-v1}~\cite{li2022devil}. These methods are all model-agnostic SGG debiasing strategies. For fair comparisons, we also reported their performance on the Motifs and VCTree baselines. 2) \textbf{KERN}~\cite{chen2019knowledge}, \textbf{G-RCNN}~\cite{yang2018graph}, \textbf{MSDN}~\cite{li2017scene}, \textbf{BGNN}~\cite{li2021bipartite}, and \textbf{DT2-ACBS}~\cite{desai2021learning}. These methods are specifically designed for SGG models. All results are reported in TABLE~\ref{tab:compare_with_sota}. Besides, we reported the R@100 metric of each predicate category group under the PredCls setting in TABLE~\ref{table:group} and presented the R@100 across all predicates in appendix. Additionally, a comparison of parameters and computational costs is provided in the appendix.

\textbf{Results.} From the results in TABLE~\ref{tab:compare_with_sota} and TABLE~\ref{table:group}, we have the following observations: 1) Compared to the two strong baselines (\ie, Motifs and VCTree), our NICE can consistently improve model performance on metric mR@K over all three tasks (\eg, 5.9\% $\sim$ 14.3\% and 3.7\% $\sim$ 14.7\% absolute gains on metric mR@100 over Motifs and VCTree, respectively). 2) Compared to the state-of-the-art model-agnostic debiasing strategies (\ie, NICE-v1~\cite{li2022devil}),
NICE can achieve comparable performance with NICE-v1 on m@K and better performance on Mean (\eg, 44.3\% \vs 43.6\% on metric Mean under PredCls over Motifs). This proves that our improvements to the NSC module do make the model more robust to all predicates compared to the original NICE-v1. 3) NICEST achieves the highest performance on the Mean metric across the three tasks (\eg, 45.3\% on Mean under PredCls over Motifs). It proves that NIST can realize a better trade-off between accuracy among different predicate categories. 4) NICEST is capable of achieving a better trade-off between head and tail predicates (\eg, 44.7\% \vs 41.7\% on Avg).

\subsubsection{Performance on VG-OOD}

\textbf{Settings.} Similarly, for the VG-OOD dataset, we also conducted NICE and NIST on the widely utilized baselines: \textbf{Motifs}~\cite{zellers2018neural} and \textbf{VCTree}~\cite{tang2019learning}. At the same time, we compared the most classic method (\ie, \textbf{TDE}~\cite{tang2020unbiased}) and current the state-of-the-art method (\ie, \textbf{NICE-v1}~\cite{li2022devil}) of model-agnostic unbiased SGG to this dataset in TABLE~\ref{tab:compare_with_sota_new_split}. In addition, we visualized the performance comparison of R@100 across all predicates and the distribution of R@100 and mR@100 in the appendix.

\textbf{Results.} From the results in TABLE~\ref{tab:compare_with_sota_new_split}, we can observe that: 1) Compared with the two common baselines (\ie, Motifs and VCTree), the mR@K of our NICE has been significantly improved in all three tasks (\eg, 4.4\% $\sim$ 17.0\% and 4.7\% $\sim$ 17.0\% absolute gains on metric mR@100 over Motifs and VCTree, respectively). 2) The NICE with an improved NSC module can outperform the original NICE-v1~\cite{li2022devil} in the comprehensive performance of the Mean metric over three tasks, which proves that the improved NSC module is indeed more robust to all predicates. 3) NICEST is designed to be independent of the frequency bias in the dataset. In terms of the VG-OOD dataset, NIST can still achieve the best performance in the Mean metric, showcasing its robust generalization capabilities and its capacity to maintain balanced performance across various predicates.

\begin{table}[!t]
  \renewcommand\arraystretch{1.0}
  \setlength\tabcolsep{3.5pt}
    \centering
    \caption{Performance (\%) on GQA dataset.}
    \vspace{-1em}
    \resizebox{0.48\textwidth}{!}{
        \renewcommand\arraystretch{1.1}
    \begin{tabular}{r||ccc}
		\hline\thickhline
        \rowcolor{mygray-bg}
         & \multicolumn{3}{c}{PredCls} \\
        \rowcolor{mygray-bg}
        \multicolumn{1}{r||}{\multirow{-2}{*}{Models~~~}}& \footnotesize{mR@50/100} & \footnotesize{R@50/100}  & \footnotesize{Mean} \\
        \hline
        \hline
        \subt{84}{22}{Motifs~\cite{zellers2018neural}}{$_{\textit{CVPR'18}}$} & 13.9 / 14.7 & 65.1 / 66.8 & 40.1 \\
        \subt{84}{22}{Motifs+NICE-v1~\cite{li2022devil}}{$_{\textit{CVPR'22}}$} & 24.8 / 27.4 & 55.8 / 58.5 & 41.6 \\
        \subt{84}{22}{\textbf{Motifs+NICE}}{~(\textbf{Ours})} & 24.5 / 27.3 & 57.1 / 59.1 & 42.0 \\
        \subt{84}{22}{\textbf{Motifs+NICEST}}{~(\textbf{Ours})} & 24.0 / 26.4 & 57.7 / 60.6 & 42.2 \\
        \hline
        \subt{84}{22}{VCTree~\cite{tang2019learning}}{$_{\textit{CVPR'19}}$} & 14.4 / 15.3 & 65.7 / 67.3 & 40.7 \\
        \subt{84}{22}{VCTree+NICE-v1~\cite{li2022devil}}{$_{\textit{CVPR'22}}$} &  24.8 / 27.2 & 56.3 / 59.2 & 41.9 \\
        \subt{84}{22}{\textbf{VCTree+NICE}}{~(\textbf{Ours})} & 25.4 / 27.9 & 56.0 / 58.8 &	42.0 \\
        \subt{84}{22}{\textbf{VCTree+NICEST}}{~(\textbf{Ours})} & 24.2 / 26.8 & 58.1 / 61.0 &	42.5 \\
        \specialrule{0.05em}{0pt}{0pt}
        \hline
    \end{tabular}
    }
    \label{tab:gqa}
\end{table}

\subsubsection{Performance on GQA}

\textbf{Settings.} For the GQA~\cite{hudson2019gqa} dataset, we assessed the performance of the following models: Motifs~\cite{zellers2018neural}, VCTree~\cite{tang2019learning}, and their combinations with NICE-v1~\cite{li2022devil}, NICE, and NICEST. The evaluation was conducted on the PredCls task. Moreover, we conducted statistical tests for comparisons on three datasets in the appendix.

\textbf{Results.} The results on the GQA dataset are summarized in TABLE~\ref{tab:gqa}. From the table, we have the following observations: 1) The incorporation of NICE and NICEST can significantly improve the mR@K scores of the two strong baselines (\eg, 12.6\% 
 and 12.6\% absolute gains on metric mR@100 over Motifs and VCTree, respectively). 2) Our NICE and NICEST are able to maintain a high and comparable R@K on the large GQA dataset. 3) NICEST achieves the highest performance on the metric Mean (\eg, 42.2\% \vs 40.1\% on metric Mean over Motifs). This demonstrates the universality and effectiveness of our method in enhancing the performance of these models.

\subsubsection{Comparisons with Label Correction and Smoothing}
\textbf{Settings.} We compared our methods (NICE and NICEST) with label correction~\cite{wang2021proselflc} and smoothing~\cite{muller2019does} methods on the PredCls task using the Motifs~\cite{zellers2018neural} model as the baseline.

\textbf{Results.}
As shown in TABLE~\ref{tab:lc}, it is evident that both NICE and NICEST achieve the highest performance in terms of the Mean metric (\eg, 44.3\% and 45.3\% under PredCls over Motifs, respectively). This far exceeds the performance of using only label correction or label smoothing methods (\eg, 3.9\% $\sim$ 4.7\% absolute gains on metric Mean). The effectiveness of our proposed methods may be attributed to the ability to identify noisy samples for each predicate, as they operate on the entire sample set and exhibit increased sensitivity towards long-tailed datasets.

\subsection{Ablation Studies on NICE}

\subsubsection{Ablation Studies on Neg-NSD}

\textbf{Hyperparameter Settings of Neg-NSD.} The hyperparameter in Neg-NSD is the threshold $\theta$ for the confidence score (\cf, Eq.~\eqref{eq:1}). Particularly, when the threshold $\theta$ for one category is set to $100\%$, which means we never assign this category as pseudo labels. Without loss of generality, we choose three representative hyperparameter settings, \ie, we mine missing annotated triplets on 1) all predicate categories, or 2) only body and tail categories, or 3) only tail categories. The thresholds $\theta$ for the corresponding head, body, and tail categories were set as $95\%$, $90\%$, and $60\%$, respectively. To disentangle the influence of the other two modules (\ie, Pos-NSD and NSC), we directly use the outputs of Neg-NSD and pristine positive samples for SGG training.

\textbf{Results.} From the results in TABLE~\ref{tab:neg_nsd}, we can observe: 1) Different threshold settings have a slight influence on the mR@K metrics, but \shr{a relatively more pronounced effect} on the R@K metrics. 2) The model gains the best performance when only mining the missing tail predicates in Neg-NSD.

\begin{table}[!t]
  \renewcommand\arraystretch{1.0}
  \setlength\tabcolsep{1pt}
    \centering
    \caption{Comparison with label correction and smoothing methods. The baseline model is Motifs~\cite{zellers2018neural}.}
    \vspace{-1em}
    \resizebox{0.48\textwidth}{!}{
        \renewcommand\arraystretch{1.1}
    \begin{tabular}{r||ccc}
	\hline\thickhline
        \rowcolor{mygray-bg}
         & \multicolumn{3}{c}{PredCls} \\
         \rowcolor{mygray-bg}
        \multicolumn{1}{r||}{\multirow{-2}{*}{Models~~~}}& \footnotesize{mR@50/100} & \footnotesize{R@50/100}  & \footnotesize{Mean} \\
        \hline  
        \hline
        \subt{98}{22}{Motifs~\cite{zellers2018neural}}{$_{\textit{CVPR'18}}$} & 16.5 / 17.8 & 65.5 / 67.2 & 41.8 \\
        \subt{98}{22}{Motifs+Label Correction~\cite{wang2021proselflc}}{$_{\textit{CVPR'21}}$} & 15.6 / 16.9 & 65.7 / 67.5 & 41.4 \\
        \subt{98}{22}{Motifs+Label Smoothing~\cite{muller2019does}}{$_{\textit{NIPS'19}}$} & 17.3 / 19.2 & 61.7 / 64.1 & 40.6 \\
        \subt{98}{22}{Motifs+NICE-v1~\cite{li2022devil}}{$_{\textit{CVPR'22}}$} & 29.9 / 32.3 &55.1 / 57.2 & 43.6 \\
        \subt{98}{22}{\textbf{Motifs+NICE}}{~(\textbf{Ours})} & 30.0 / 32.1 & 56.6 / 58.6 & 44.3 \\
        \subt{98}{22}{\textbf{Motifs+NICEST}}{~(\textbf{Ours})} & 29.5 / 31.6 & 59.1 / 61.0 & 45.3 \\
        \specialrule{0.05em}{0pt}{0pt}
        \hline
    \end{tabular}
    }
    \label{tab:lc}
\end{table}

\begin{table}[!t]
\renewcommand\arraystretch{1.0}
  \setlength\tabcolsep{6pt}
    \centering
    \caption{Recall@100 of each group,\ie, head, body, and tail, under PredCls setting. Avg is the average of the three groups.}
    \vspace{-1em}
    \resizebox{0.48\textwidth}{!}{
    \renewcommand\arraystretch{1.1}
    \begin{tabular}{r||cccc}
    \hline\thickhline
    \rowcolor{mygray-bg} & \multicolumn{4}{c}{PredCls} \\
    \rowcolor{mygray-bg}
    \multicolumn{1}{r||}{\multirow{-2}{*}{Models~~~}}
    & {Head}  & {Body}  & {Tail}  & {Avg} \\
    \hline
    \hline
    \subt{84}{22}{Motifs~\cite{zellers2018neural}}{$_{\textit{CVPR'18}}$} & 80.5  & 29.3  & 5.7   & 38.5 \\
    \subt{84}{22}{Motifs+NICE-v1~\cite{li2022devil}}{$_{\textit{CVPR'22}}$} & 67.9  & 33.1  & 24.2  & 41.7 \\
    \subt{84}{22}{\textbf{Motifs+NICE}}{~(\textbf{Ours})} & 69.6  & 36.9  & 26.0  & 44.2 \\
    \subt{84}{22}{\textbf{Motifs+NICEST}}{~(\textbf{Ours})} & 72.9  & 35.7  & 25.4  & 44.7 \\
    \specialrule{0.05em}{0pt}{0pt}
    \hline
    \end{tabular}}
    \label{table:group}
\end{table}

\textbf{Effectiveness of Neg-NSD.} We evaluated the effectiveness of the Neg-NSD by using the same refined samples by Neg-NSD and pristine positive samples for SGG training.

\textbf{Results.} All results are reported in TABLE~\ref{tab:component}. Compared to the baseline model (\#~1), the Neg-PSD module (\#~2) can significantly improve model performance on mR@K metrics (\eg, $25.2\%$ \vs $17.8\%$ in mR@100), which proves that these harvested ``positive" samples (noisy negative samples with pseudo labels) are indeed beneficial for SGG training.

\begin{table*}[!htbp]
    \caption{Ablation studies on the influence of different hyperparameters of each component of NICE. Motifs~\cite{zellers2018neural} is the baseline model which are used in all experiments.}
    \vspace{-1em}
    \begin{minipage}{0.39\linewidth}
    \centering
    \renewcommand\arraystretch{1.0}
        \setlength\tabcolsep{1.9pt}
        \captionsetup{type=figure} 
        \captionsetup[subfloat]{position=bottom}
        \subfloat[Ablation studies (\%) on each component of NICE. ``\#" is the line number.]{
         \resizebox{\textwidth}{!}{
        \renewcommand\arraystretch{1.0}
        \begin{tabular}{l|ccc||cccc}
        \hline\thickhline
        \rowcolor{mygray-bg}
        & \multicolumn{3}{c||}{Components} & \multicolumn{3}{c}{PredCls} \\
        \rowcolor{mygray-bg}
        \# & N-NSD & {P-NSD} & {NSC}     & {mR@50/100} & {R@50/100}  & {Mean} \\
        \hline
        \hline
        1 & \XSolidBrush & \XSolidBrush & \XSolidBrush  & 16.5 / 17.8    & 65.5 / 67.2  & 41.8 \\
        2 & \Checkmark & \XSolidBrush & \XSolidBrush  & 23.3 / 25.2  &  62.3 / 64.5  & 43.8 \\
        3 & \XSolidBrush & \Checkmark & \XSolidBrush & 20.3 / 22.0    & 57.6 / 59.2  & 39.8 \\
        4 & \Checkmark & \Checkmark & \XSolidBrush & 20.4 / 23.4 & 56.7 / 61.4  & 40.5 \\
        5 & \XSolidBrush & \Checkmark & \Checkmark & 23.3 / 25.2  & 59.6 / 61.3  & 42.4 \\
        6 & \Checkmark & \Checkmark & \Checkmark & 30.0 / 32.1  & 56.6 / 58.6  & 44.3 \\
        \specialrule{0.05em}{0pt}{0pt}
        \hline
        \end{tabular}
        \label{tab:component}
        }}
        \vspace{-1em}
        \renewcommand\arraystretch{1.0}
        \setlength\tabcolsep{4.0pt}
        \subfloat[Ablation studies (\%) on different cutoff distances $d_c$ in Pos-NSD for head, body and tail predicates, respectively.]{
             \resizebox{\textwidth}{!}{
            \renewcommand\arraystretch{1.05}
            \begin{tabular}{ccc||ccc}
            \hline\thickhline
		\rowcolor{mygray-bg}
            \multicolumn{3}{c||}{Size of $d_c$} & \multicolumn{3}{c}{PredCls} \\
            \rowcolor{mygray-bg}
            head & body & tail  & mR@50/100   & R@50/100  & Mean \\
            \hline
            \hline
            $\mathcal{L}$ & $\mathcal{L}$ & $\mathcal{L}$ & 19.9 / 21.5  & 64.0 / 65.7  & 42.8 \\
            $\mathcal{M}$ & $\mathcal{M}$ & $\mathcal{M}$ & 21.0 / 22.8  & 62.2 / 64.0  & 42.5 \\
            $\mathcal{S}$ & $\mathcal{S}$ & $\mathcal{S}$ & 21.6 / 23.4  & 61.1 / 62.8  & 42.2 \\
            $\mathcal{S}$ & $\mathcal{M}$ & $\mathcal{L}$ & 23.3 / 25.2  & 59.6 / 61.3  & 42.4 \\
            $\mathcal{L}$ & $\mathcal{M}$ & $\mathcal{S}$ & 18.6 / 20.1  & 64.4 / 66.1  & 42.3 \\
            \specialrule{0.05em}{0pt}{0pt}
            \hline
         \end{tabular}
         \label{tab:pos_nsd}
         }}
        \end{minipage} 
        \hfill
        \begin{minipage}[!r]{0.59\linewidth}
        \captionsetup{type=figure}
        \captionsetup[subfloat]{position=bottom}
        \renewcommand\arraystretch{1.2}
        \setlength\tabcolsep{1.6pt}
        \vspace{-1.2em}
        \subfloat[Ablation studies (\%) on mining missing annotated triplets in different categories in Neg-NSD.]{
         \resizebox{0.55\textwidth}{!}{
        \renewcommand\arraystretch{1.1}
        \begin{tabular}{ccc||ccc}
            \hline\thickhline
            \rowcolor{mygray-bg}
            \multicolumn{3}{c||}{Categories} & \multicolumn{3}{c}{PredCls} \\
            \rowcolor{mygray-bg}
            head & body & tail & mR@50/100 & R@50/100  & Mean \\
            \hline
            \hline
            \Checkmark & \Checkmark & \Checkmark & 22.2 / 25.8  & 58.5 / 62.5  & 42.3 \\
            \XSolidBrush & \Checkmark & \Checkmark & 23.7 / 26.2  & 59.5 / 62.3  & 42.9 \\
            \XSolidBrush & \XSolidBrush & \Checkmark & 23.3 / 25.2  & 62.3 / 64.5  & 43.8 \\
            \specialrule{0.05em}{0pt}{0pt}
            \hline
        \end{tabular}
        \label{tab:neg_nsd}
        }}
         \renewcommand\arraystretch{1.2}
        \setlength\tabcolsep{2.9pt}
        \subfloat[Ablation studies (\%) on different $K$ in wKNN.]{
         \resizebox{0.44\textwidth}{!}{
         \renewcommand\arraystretch{1.2}
        \begin{tabular}{c||ccc}
    	\hline\thickhline
            \rowcolor{mygray-bg}
            & \multicolumn{3}{c}{PredCls} \\
            \rowcolor{mygray-bg}
            \multicolumn{1}{c||}{\multirow{-2}{*}{$K$}} 
            & mR@50/100 & R@50/100  & Mean \\
            \hline
            \hline
            1 & 23.3 / 25.0 & 59.3 / 60.9 & 42.1 \\
            3 & 23.3 / 25.2 & 59.6 / 61.3 & 42.4 \\
            5 & 22.9 / 24.7 & 59.8 / 61.5 & 42.2 \\
            \specialrule{0.05em}{0pt}{0pt}
            \hline
        \end{tabular}
        \label{tab:k_nsc}
        }}

    \includegraphics[width=0.98\linewidth]{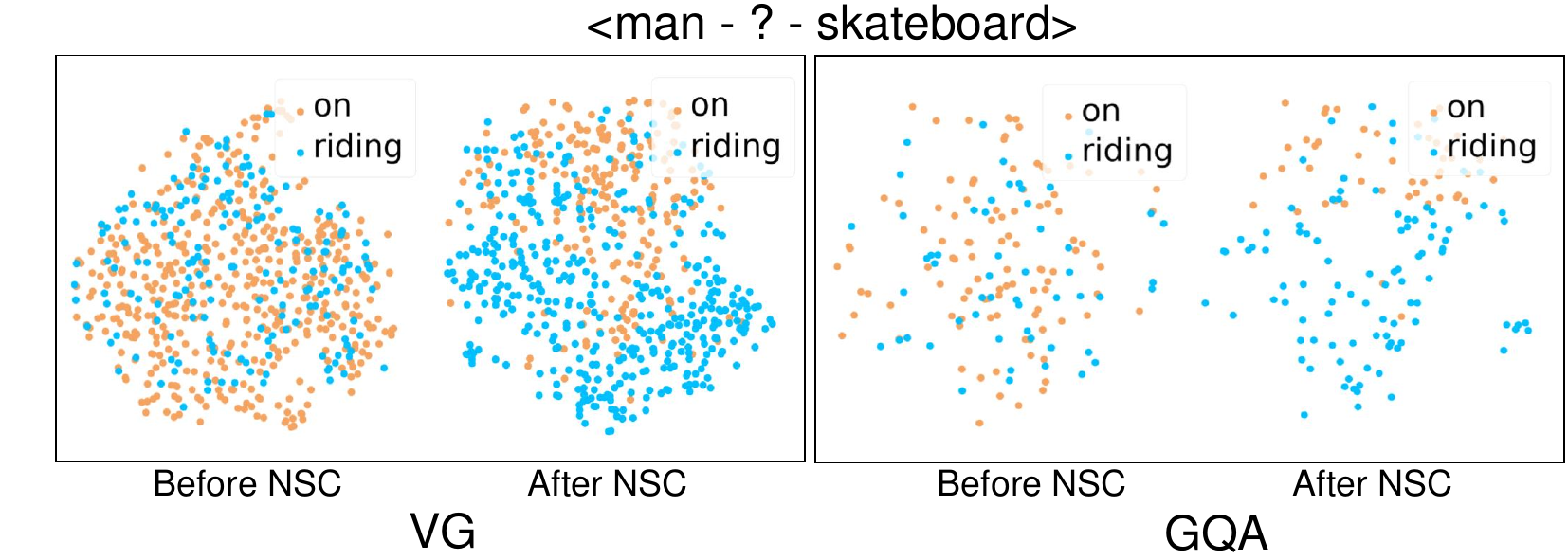}
        \vspace{-1.0em}
          \caption{The t-SNE visualization of randomly sampled instances of \texttt{man}-\texttt{on}/\texttt{riding}-\texttt{skateboard} triplet categories on feature space before and after NSC.}
          \label{fig:t-sne}
    \end{minipage}
    \label{tab:abla}
\end{table*}

\addtolength{\tabcolsep}{2pt}
\begin{table*}[htbp]
  \renewcommand\arraystretch{1.2}
  \setlength\tabcolsep{8pt}
  \centering
  \caption{Ablation studies (\%) on the combination of different teacher models in NIST.}
  \vspace{-1em}
  \resizebox{\textwidth}{!}{
  \renewcommand\arraystretch{1.1}
  \begin{tabular}{cc|c|ccc|c|ccc}
    \hline\thickhline
    \multicolumn{2}{c|}{\cellcolor{mygray-bg}{Teacher}} & \cellcolor{mygray-bg} & \multicolumn{3}{c|}{\cellcolor{mygray-bg}{PredCls}} & \cellcolor{mygray-bg} & \multicolumn{3}{c}{\cellcolor{mygray-bg}{PredCls}} \\
    \rowcolor{mygray-bg}
        $T^{Head}$    & $T^{Tail}$    & \multirow{-2}{*}{Dataset} & mR@50/100 & R@50/100 & Mean  & \multirow{-2}{*}{Dataset} & mR@50/100 & R@50/100 & Mean \\
    \hline\hline
    \multirow{3}[2]{*}{   Motifs   } & -     &   \multirow{4}[6]{*}{VG}    &  14.0	/ 15.3	& 65.2 / 67.1 & 40.4 &   \multirow{4}[6]{*}{VG-OOD}    & 11.3 / 11.9 & 57.8 / 58.4 & 34.9  \\
          &  Motifs-TDE   & & 25.7 / 29.7	& 45.7 / 51.3 & 38.1 
       &     
          & 20.4 / 22.5 & 56.8 / 60.0 & 39.9 \\
         & Motifs-NICE  &       &   29.5 / 31.6 &  59.1 / 61.0 & 45.3 &        &  26.4 / 27.1 & 54.8 / 55.5 & 41.0 \\
\cline{1-2}\cline{4-6}\cline{8-10}    \multirow{3}[2]{*}{   VCTree 
  } & -     &       &   17.1 / 18.4 & 65.9 / 67.6 & 42.3    &   & 12.1	/ 12.5 & 58.2	/ 58.8 & 35.4 \\
          & VCTree-TDE & & 26.9 / 30.5 & 45.6 / 49.7 & 38.2 
 & & 22.4 / 25.7 & 54.2/ 57.8 & 40.0  \\
          & VCTree-NICE  &       &       30.6 / 32.9 & 59.1 / 60.9 & 45.9  &       & 26.6 / 27.4 & 54.3 / 55.0 & 40.8  \\
          
    \specialrule{0.05em}{0pt}{0pt}
    \hline
    \end{tabular}%
    }
  \label{tab:tradeoff}%
\end{table*}%
\addtolength{\tabcolsep}{-2pt}

\subsubsection{Ablation Studies on Pos-NSD}

\noindent\textbf{Hyperparameter Settings of Pos-NSD.} The hyperparameter in Pos-NSD is the cutoff distance $d_c$ ranked at $\alpha$\% for different categories (\cf, Eq.~\eqref{eq:5}). As mentioned in Sec.~\ref{sec:3.2}, different $d_c$ directly \shr{impacts} the clustering results of each predicate category, and smaller $d_c$ is more suitable for predicates with multiple semantic meanings. Therefore, without loss of generality, we choose five typical settings for different categories, including ($\mathcal{L}, \mathcal{L}, \mathcal{L}$),
($\mathcal{M}, \mathcal{M}, \mathcal{M}$), and so on. \shr{In our experiments,} $\mathcal{L}$, $\mathcal{M}$, $\mathcal{S}$ denote \shr{that $d_c$ is} large, medium, and small, respectively. Their corresponding $\alpha\%$ values were set to $50.0\%$, $25.0\%$ and $12.5\%$. Similarly, we disentangle the influence of Neg-NSD and use pristine negative and refined positive samples (outputs of NSC) for SGG training. The results are shown in TABLE~\ref{tab:pos_nsd}.

\textbf{Results.} When the cutoff distance $d_c$ for all predicate categories is set to large or small or from large to small, the model achieves relatively worse results. These results are also consistent with our expectation, \ie, for the predicate categories with multiple semantic meanings (head categories), small $d_c$ is better for noisy sample detection. Instead, for predicate categories with a unique semantic meaning (tail categories), a larger $d_c$ is better. Thus, we utilize the ($\mathcal{S}, \mathcal{M}, \mathcal{L}$) setting for all experiments.

\textbf{Effectiveness of Pos-NSD.} We evaluated the effectiveness of Pos-NSD by using either clean positive samples detected from $\bm{\mathcal{T}}^+$ with pristine negative samples or from the new positive samples set ${\widetilde {\bm{\mathcal{T}}}^+}$ for SGG training.

\textbf{Results.} As shown in TABLE~\ref{tab:component} (\# 3), the single Pos-NSD component can still improve mR@K metrics using much fewer positive samples. Besides, the baseline can also be exceeded on mR@K (\# 4) after Neg-NSD and Pos-NSD alone with fewer training samples. It proves that the clean subset divided by Pos-NSD is better for SGG training, and numerous noisy positive samples actually hurt performance.

\subsubsection{Ablation Studies on NSC} 
\noindent\textbf{Hyperparameter of NSC.} The hyperparameter of NSC is the $K$ in wKNN. We investigated $K=\{1,3,5\}$. All results are reported in TABLE~\ref{tab:k_nsc}. From the results, we can observe that the SGG performance is robust to different $K$. To better make a trade-off of different metrics, we set $K$ to 3.

\textbf{Effectiveness of NSC.} Based on Pos-NSD, NSC replaces the original labels of the noisy positive samples with cleaner and more robust soft labels. As shown in TABLE~\ref{tab:component}, NSC markedly improves SGG performance on both mR@K and R@K (\#~5 \vs \#~3), \ie, NSC reassigns better labels to noisy samples.

\textbf{Results.} Compared with the NSC in NICE-v1~\cite{li2022devil}, both mR@K and R@K of PredCls are improved by our refined NSC method (33.1\% \vs 33.0\% in mR@100 and 58.3\% \vs 56.9\% in R@100, based on VCTree). It proves that our approach is relatively improved across all predicate categories.

\addtolength{\tabcolsep}{2pt}
\begin{table}[!t]
  \renewcommand\arraystretch{1.2}
  \setlength\tabcolsep{6pt}
  \centering
  \vspace{-0.5em}
  \caption{Ablation studies (\%) on different weighting methods of different teacher models in NIST.}
  \vspace{-1em}
  \resizebox{0.48\textwidth}{!}{
\renewcommand\arraystretch{1.1}
    \begin{tabular}{ccc||ccc}
    \hline\thickhline
    \rowcolor{mygray-bg}
      \multicolumn{3}{c||}{Weighting Type} & \multicolumn{3}{c}{PredCls} \\
      \rowcolor{mygray-bg}
      \multicolumn{1}{c}{Fixed} & \multicolumn{1}{c}{Adapt} & \multicolumn{1}{c||}{Group} & \multicolumn{1}{c}{mR@50/100} & \multicolumn{1}{c}{R@50/100} & \multicolumn{1}{c}{Mean} \\
    \hline\hline
     \Checkmark &  &  & 28.9 / 31.1 & 57.9 / 59.8 & 44.4   \\
       &  \Checkmark &   &  30.4 / 32.6 & 57.4 / 59.4 & 45.0 \\
   &   & \Checkmark &  29.5 / 31.6 &  59.1 / 61.0 & 45.3 \\
   \specialrule{0.05em}{0pt}{0pt}
    \hline
    \end{tabular}%
    }
  \label{tab:weight}%
  \vspace{-0.5em}
\end{table}%
\addtolength{\tabcolsep}{-2pt}

\textbf{T-SNE Visualizations.} We visualized the t-SNE distributions of features of $\langle$\texttt{man}-\texttt{on/riding}-\texttt{skateboard}$\rangle$ on the VG and GQA datasets\shr{, comparing} the feature space before and after the NSC in Fig.~\ref{fig:t-sne}. The former is a pair of predicates of different granularity (common-prone), while the latter is a pair of predicates of the same granularity (synonym-random). As shown in Fig.~\ref{fig:t-sne}, NSC can help to alleviate the inconsistency of ground-truths, \ie, similar visual patterns always have more consistent ground-truth predicate annotations, which is beneficial for SGG training. 

\subsection{Ablation Studies on NIST} \label{nist}

\begin{figure*}[!t]
    \centering
    \includegraphics[width=\linewidth]{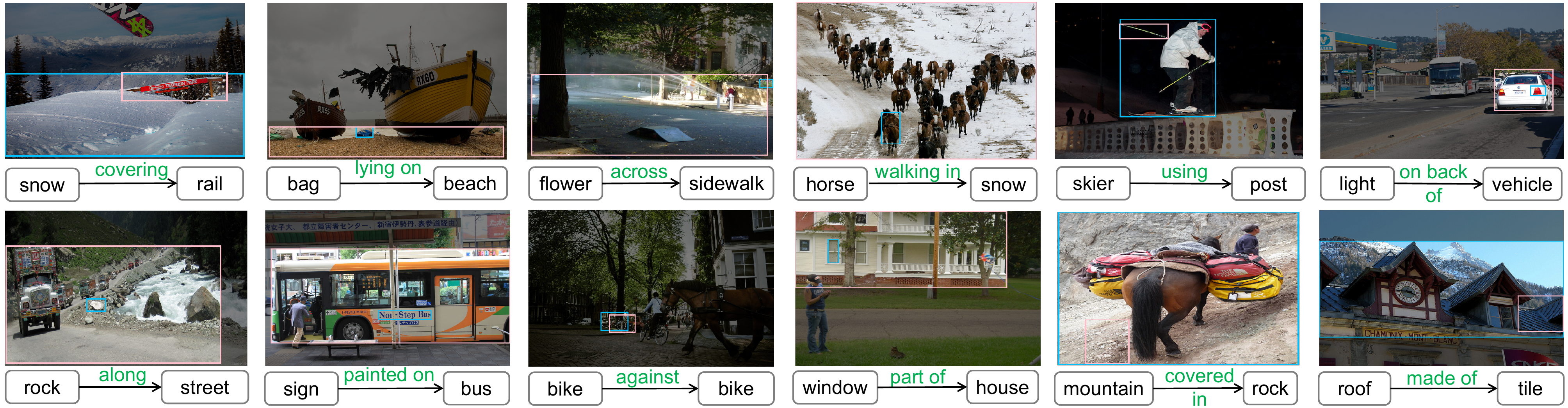}
    \vspace{-2.0em}
    \caption{The examples of the triplets detected by Negative Noisy Sample Detection that do not appear in the training set. }
  \label{fig:nsd_example}
\end{figure*}

\begin{figure*}[!t]
    \centering
        \vspace{-0.5em}
    \includegraphics[width=\linewidth]{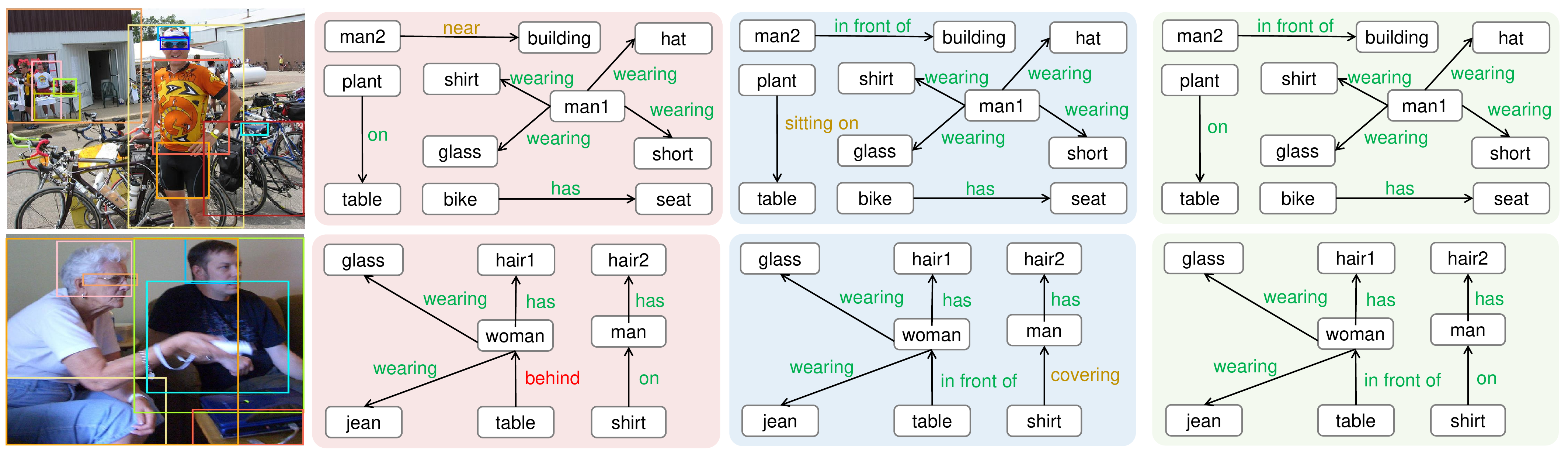}
    \vspace{-2.0em}
    \caption{Scene graphs generated by Motifs (left) and Motifs+NICE (Middle), Motifs+NICEST (right) on PredCls. \textcolor{red}{\textbf{Red}} predicates are error (\ie, not GT and unreasonable), \textcolor{mygreen}{\textbf{Green}} predicates are correct (\ie, GT), and \textcolor{brown}{\textbf{Brown}} predicates are reasonable (not in GT but reasonable). Only detected boxes overlapping with GT are shown.}
    \label{fig:vis}
    \vspace{-1em}
\end{figure*}

\noindent\textbf{Hyperparameter of NIST.} The hyperparameter of NIST is the weight of two teachers. We evaluated the performance of the three weighted methods in NIST. Among them, ``fixed" means that the weights of two teacher models are constant 0.5, ``adapt" means that two teacher models adopt adaptive trade-off strategies (\cf  Eq.~\ref{eq:tradeoff}), and ``group" means that two teacher models adopt different adaptive weighting strategies for the head, body, and tail category groups respectively.
All results are reported in TABLE~\ref{tab:weight}.

\textbf{Results.} As displayed in TABLE~\ref{tab:weight}, our proposed group weighting strategy achieves the best performance on the Mean metric, proving the effectiveness of the grouping weighting.

\textbf{Effectiveness of NIST.} 
To prove the generality of NIST, we used the two most common baseline models (\ie, \textbf{Motifs}~\cite{zellers2018neural} and \textbf{VCTree}~\cite{tang2019learning}) as teachers \shr{biased} towards head predicates (\ie, high Recall), and two models trained using unbiased methods (\ie, Motifs-TDE~\cite{tang2020unbiased} and Motifs-NICE) as teachers \shr{biased} towards the tail (\ie, high mean Recall). The results of the combination of different teacher models in NIST are reported in TABLE~\ref{tab:tradeoff}.

\textbf{Results.} In TABLE~\ref{tab:compare_with_sota},  TABLE~\ref{tab:compare_with_sota_new_split}, and Tabel~\ref{tab:tradeoff}, we can find that the model trained by NIST can effectively improve the Mean performance across all the baseline models and \shr{datasets} (\eg, 36.9\% \vs 38.1\% and 37.7\% \vs 39.9\% based on Motifs for TDE on VG and VG-OOD, respectively). Consistent improvement across all models and datasets demonstrates the universality of NIST.

\subsection{Visualization}

\subsubsection{Qualitative Results of Neg-NSD}\label{sec:b} 
In Fig.~\ref{fig:nsd_example}, we visualized some ``unseen" visual relation triplet categories mined by Neg-NSD that never appear in the original VG dataset. Some of these triplets are easily overlooked by the annotators, such as the relation \texttt{against} between \texttt{bike} and \texttt{bike}, or the relation \texttt{along} between \texttt{rock} and \texttt{street}. These harvested new visual relation triplet categories increase both the number and diversity of samples in tail categories.

\subsubsection{Qualitative Results of NICE}
Fig.~\ref{fig:vis} demonstrates some qualitative results generated by Motifs, Motifs+NICE, and Motifs+NICEST. From Fig.~\ref{fig:vis}, we can observe that Motifs tends to predict coarse-grained (\ie, head) predicates, such as \texttt{near}, while Motifs+NICE tends to predict fine-grained (\ie, tail) predicates, such as \texttt{sitting on} and \texttt{covering}. The Motifs+NICEST can obtain trade-off predicates that bring predictions closer to the ground-truth.


\section{Conclusions and Limitations}

In this paper, we argued that two plausible assumptions about the ground-truth annotations are inapplicable to existing SGG datasets. To this end, we reformulated SGG as a noisy label learning problem and proposed a novel model-agnostic noisy label correction and sample training strategy: NICEST. It is composed of NICE and NIST, which solves the problem of noisy label learning by generating high-quality samples and efficient training, respectively. NICE can not only detect the noisy samples, but also reassign robust soft predicate labels to them. NIST compensates for the tail-biased training defect of NICE by adopting a multi-teacher distillation strategy to enable the model to learn unbiased fusion knowledge. We re-organized the VG dataset to create VG-OOD for a better evaluation of the model generalization capability. Extensive experiments on VG, VG-OOD and GQA datasets prove the effectiveness of each component of NICEST.

\textbf{Limitations} The use of Multi-Teacher Knowledge Distillation in NICEST may introduce minor computational overhead. In the case of online training, it requires twice the GPU space compared to the baseline. However, with offline training, there is no additional GPU consumption overhead. Additionally, since certain hyperparameters have varying impacts on different predicate categories, it becomes challenging to achieve the optimal trade-off between different evaluation metrics.

\section*{Acknowledgments}
This work was supported by the National Key Research \& Development Project of China (2021ZD0110700), the National Natural Science Foundation of China (62337001) and the Fundamental Research Funds for the Central Universities. Long Chen was supported by HKUST Special Support for Young Faculty (F0927) and HKUST Sports Science and Technology Research Grant (SSTRG24EG04).
\bibliographystyle{IEEEtran}
\bibliography{IEEEfull.bib}
\end{document}